\documentclass[10pt,twocolumn,letterpaper]{article}

\usepackage[dvipsnames,HTML]{xcolor}
\usepackage{cvpr} % To produce the CAMERA-READY version

\usepackage{algorithm}
\usepackage{algpseudocode}
\usepackage{algorithmicx}

\definecolor{cvprblue}{rgb}{0.21,0.49,0.74}
\usepackage[pagebackref,breaklinks,colorlinks,allcolors=cvprblue]{hyperref}

\usepackage{multicol}
\usepackage{multirow}

\usepackage{multirow}
\usepackage{array}
\usepackage{graphicx}
\usepackage{amsmath}    % Provides equation environments like align, gather, etc.
\usepackage{amssymb}    % Provides additional mathematical symbols
\usepackage{amsfonts}   % Provides additional fonts for math
\usepackage{makecell}

\usepackage{bm} % For bold math
\usepackage{booktabs} % For table styling
\usepackage[most]{tcolorbox}

\title{UPME: An Unsupervised Peer Review Framework for\\ Multimodal Large Language Model Evaluation}

\definecolor{LightBack}{RGB}{247,249,251}

\author{
    Qihui Zhang\textsuperscript{1,3*} \quad
    Munan Ning\textsuperscript{1*} \quad
    Zheyuan Liu\textsuperscript{1,3} \quad
    Yue Huang\textsuperscript{4} \quad
    Shuo Yang\textsuperscript{1} \\[2pt]
    Yanbo Wang\textsuperscript{1} \quad
    Jiayi Ye\textsuperscript{1} \quad
    Xiao Chen\textsuperscript{5} \quad
    Yibing Song\textsuperscript{2,3} \quad
    Li Yuan\textsuperscript{1$\dagger$} \\[10pt]
    $^1$School of Electrical and Computer Engineering, Peking University\quad
    $^2$Hupan Lab\\[2pt]
    $^3$DAMO Academy, Alibaba Group\quad
    $^4$University of Notre Dame\quad
    $^5$Tsinghua University\\
}
\vspace{-5pt}

\tcbset{
    prompt/.style={
        colback=LightBack,
    colframe=black,
    boxsep=0.5pt,
    arc=3pt,
    boxrule=1pt,
    width=\textwidth,
    enhanced,
    drop shadow={black!50!white},
    boxed title style={
        colback=white,
        colframe=black,
        boxrule=1pt,
        sharp corners,
        size=small,
        boxsep=2pt,
        left=2mm,
        right=2mm,
        fontupper=\color{black}

    },
    attach boxed title to top left={yshift=-\tcboxedtitleheight/2},
    top=3mm,
    bottom=2mm,
coltitle=black
    }
}

\tcbset{
    guideline/.style={
        colback=LightBack,
        colframe=black,
        boxsep=0.5pt,
        arc=3pt,
        boxrule=1pt,
        width=\textwidth,
        enhanced,
        drop shadow={black!50!white},
        boxed title style={
            colback=orange!20,
            colframe=black,
            boxrule=1pt,
            sharp corners,
            size=small,
            boxsep=2pt,
            left=2mm,
            right=2mm,
            fontupper=\color{black}
        },
        attach boxed title to top left={yshift=-\tcboxedtitleheight/2},
        top=3mm,
        bottom=2mm,
        coltitle=black
    }
}

\newcommand\nonumfootnote[1]{%
\begingroup%
    \renewcommand\thefootnote{}\footnote{\hspace{-3.7pt}#1}%
    \addtocounter{footnote}{-1}%
\endgroup%
}

\begin{document}
\maketitle
\begin{abstract}
Multimodal Large Language Models (MLLMs) have emerged to tackle the challenges of Visual Question Answering (VQA), sparking a new research focus on conducting objective evaluations of these models. Existing evaluation methods face limitations due to the significant human workload required to design Q\&A pairs for visual images, which inherently restricts the scale and scope of evaluations. Although automated MLLM-as-judge approaches attempt to reduce the human workload through automatic evaluations, they often introduce biases.
To address these problems, we propose an \textbf{U}nsupervised \textbf{P}eer review \textbf{M}LLM \textbf{E}valuation framework. It utilizes only image data, allowing models to automatically generate questions and conduct peer review assessments of answers from other models, effectively alleviating the reliance on human workload.
Additionally, we introduce the vision-language scoring system to mitigate the bias issues, which focuses on three aspects: \( (i)\) response correctness; \( (ii)\) visual understanding and reasoning; and \( (iii)\) image-text correlation.
Experimental results demonstrate that UPME achieves a Pearson correlation of 0.944 with human evaluations on the MMstar dataset and 0.814 on the ScienceQA dataset, indicating that our framework closely aligns with human-designed benchmarks and inherent human preferences.
\end{abstract}

\nonumfootnote{* Equal contribution \qquad $\dagger$ Corresponding author \\
\texttt{\{maskhui1003, munanning\}@gmail.com yuanli-ece@pku.edu.cn}
}
\vspace{-20pt}
\section{Introduction}
\label{sec:intro}

Recently, Large Language Models (LLMs) have made significant strides in reasoning and application capabilities~\cite{openai2023gpt4,anthropic2024claude35,wei2022chain}, enabling diverse applications such as text and code generation~\cite{achiam2023gpt, kasner2024beyond, wu2024unigen, roziere2023code}. Notably, due to the exceptional natural language understanding capabilities of LLMs, they are increasingly employed for automated evaluations, a process known as LLM-as-a-Judge~\cite{zheng2023judging}. With the integration of vision encoders \cite{pi2024perceptiongpt}, Multimodal Large Language Models (MLLMs)~\cite{liu2024visual,openai2023gpt4} can incorporate multiple modalities (e.g., text, image, and audio)~\cite{lin2023video,zhu2023languagebind} and showcase remarkable performance in downstream applications, including visual conversation \cite{cai2023low} and embodied intelligence~\cite{driess2023palm}. Furthermore, they can act as agents to interact with GUI~\cite{chen2024gui, hong2024cogagent}.

\begin{figure}[t]
  \centering
   \includegraphics[width=\linewidth]{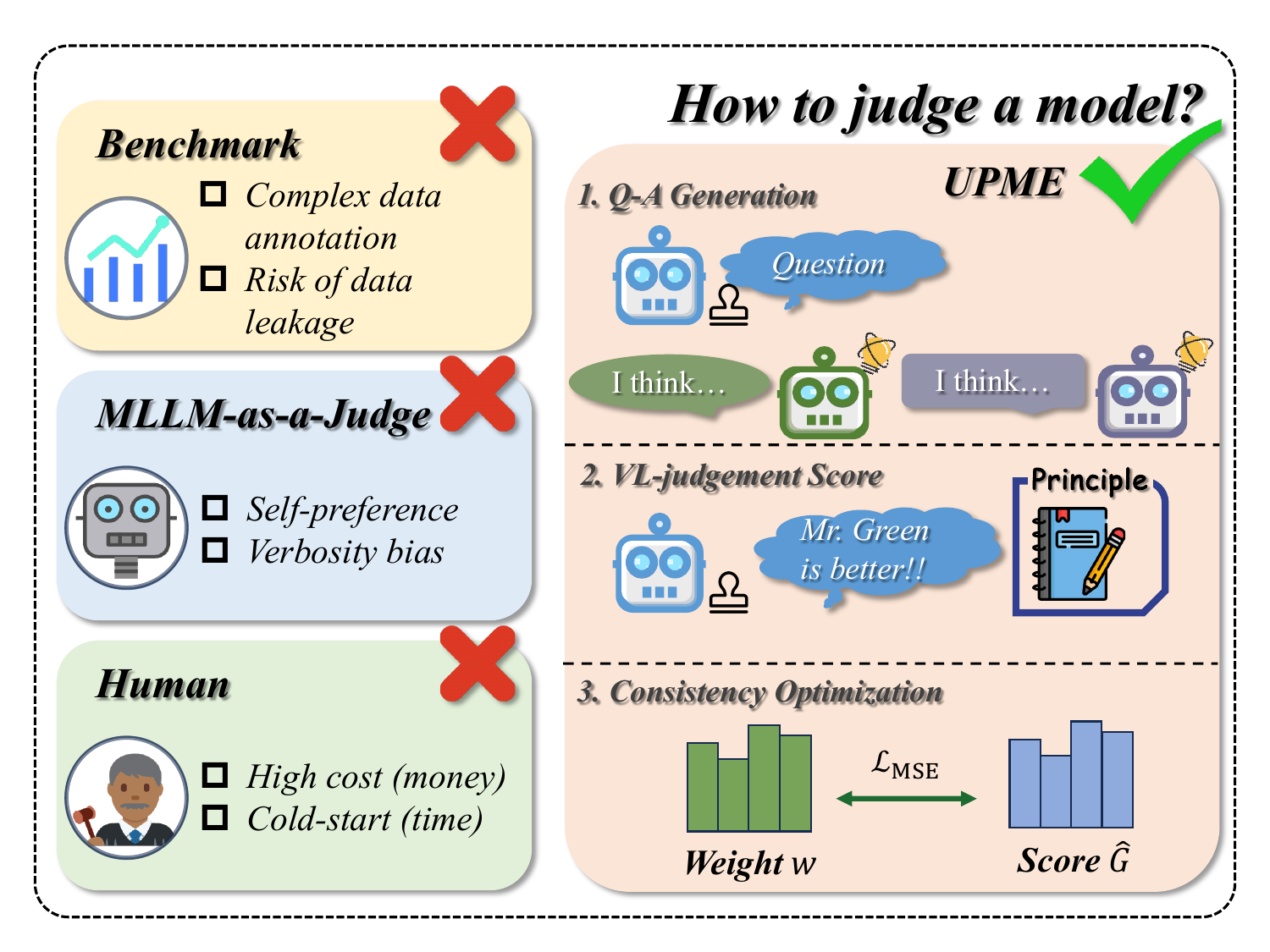}
    \vspace{-15pt}
   \caption{Existing methods for evaluating MLLMs face various challenges. Our proposed UPME framework addresses these limitations by leveraging a peer review mechanism, reducing annotation costs, and aligning closely with human judgment.}
   \label{fig:onecol}
   \vspace{-10pt}
\end{figure}

Subsequently, the challenge of conducting objective and comprehensive evaluations of MLLMs while providing users with authoritative guidelines has emerged as a key research focus. A common approach involves constructing benchmarks based on Visual Question Answering (VQA), which assess the performance of MLLMs across various dimensions by setting image-based questions and answers~\cite{lu2022learn, yu2023mm, yue2024mmmu, chen2024we}. However, most of these works heavily depend on substantial human involvement in creating Q\&A pairs for images~\cite{li2024survey, tang2020semantic}, leading to limited evaluation volume and making it insufficient for a comprehensive assessment of visual content~\cite{bao2024autobenchv}.
Another category of methods is MLLM-as-a-Judge~\cite{zheng2023judging,liu2024aligning,gao2024best,kasner2024traditionalbenchmarksanalyzingbehaviors}, which aims to reduce reliance on human annotations by allowing models to evaluate each other directly. However, these methods often still depend on human-designed questions and cannot entirely eliminate the need for human input. Additionally, a significant challenge is the presence of verbosity and self-preference biases \cite{chen2024mllm, ye2024justice}, where MLLMs tend to favor longer responses or prefer their own outputs. These biases can lead mutual evaluation mechanisms to deviate from truly understanding visual content.

To address these challenges, we propose UPME, the \textbf{U}nsupervised \textbf{P}eer review \textbf{M}LLM \textbf{E}valuation framework, which is designed to evaluate MLLMs without requiring human QA annotations, as shown in~\autoref{fig:onecol}. During each iteration, UPME selects two candidate models and a review model from the MLLM pool. The review model generates questions for a given image and evaluates the responses provided by the candidate models. This evaluation is conducted using a vision-language scoring system that assesses textual response correctness, visual understanding and reasoning, as well as image-text consistency through CLIP scores. 
Each model is initialized with a confidence weight and alternates roles as candidate and review models. Their estimated scores are calculated based on weighted evaluations provided by the review models. The framework employs dynamic weight optimization, iteratively updating model scores and refining weights through consistency optimization. Ultimately, this process generates a comprehensive and less biased estimated score list.

We selected the MMStar~\cite{chen2024we} and ScienceQA~\cite{lu2022learn} datasets as benchmarks because they encompass visual content reasoning and multimodal content across various disciplines. Experimental results show that under unsupervised settings, UPME achieved Pearson correlations of 0.944 and 0.814 with human evaluation results on these two datasets, while Spearman similarities reached 0.972 and 0.886, indicating a high degree of similarity between our approach and human-annotated QA benchmarks. Further experiments demonstrated that UPME achieved higher consistency with human assessments compared to purely peer review-based methods, effectively reducing verbosity and self-preference issues in the MLLM-as-a-Judge framework.

Overall, our contributions can be summarized as follows:
\begin{itemize}
    \item We propose the first \textbf{U}nsupervised \textbf{P}eer review \textbf{M}LLM \textbf{E}valuation (UPME) framework, addressing the challenge of reliance on human annotations in MLLM evaluation
    \item Our framework incorporates a Vision-Language Scoring System that targets visual performance and image-text association, providing a visually-focused evaluation.
    \item We conduct extensive experiments demonstrating that UPME achieves high consistency with human QA assessments and strong alignment with human preferences.
\end{itemize}
\section{Related Work}
\label{sec:RelatedWork}

\subsection{MLLM Benchmark}

MLLM benchmarks rely on human-annotated QA pairs to measure the accuracy of MLLMs in answering questions. They focus on the different capabilities of MLLMs. ScienceQA~\cite{lu2022learn} evaluate the multi-turn reasoning ability for a broad range of scientific subjects by introducing Chain of Thought (CoT) \cite{wei2022chain}; MMVet~\cite{yu2023mm} presents an LLM-based open-ended output evaluation method, enabling unified scoring across various question types and answer formats; MMMU~\cite{yue2024mmmu}, inspired by MMLU, primarily assesses model performance in complex reasoning tasks requiring college-level knowledge, covering six major disciplines and 30 fields with multimodal questions; MMStar~\cite{chen2024we} focuses on creating a benchmark requiring visual content for reasoning, rigorously selecting visual-dependent samples and detecting data leakage to ensure genuine performance gains from multimodal training; RealWorldQA~\cite{RealWorldQA} is designed to assess the ability of MLLMs to understand real-world space. Although these benchmarks propose various evaluation methods, their evaluation scope is limited by the reliance on substantial human annotation.

\begin{figure*}[!t]
    \centering
    \includegraphics[width=\linewidth]{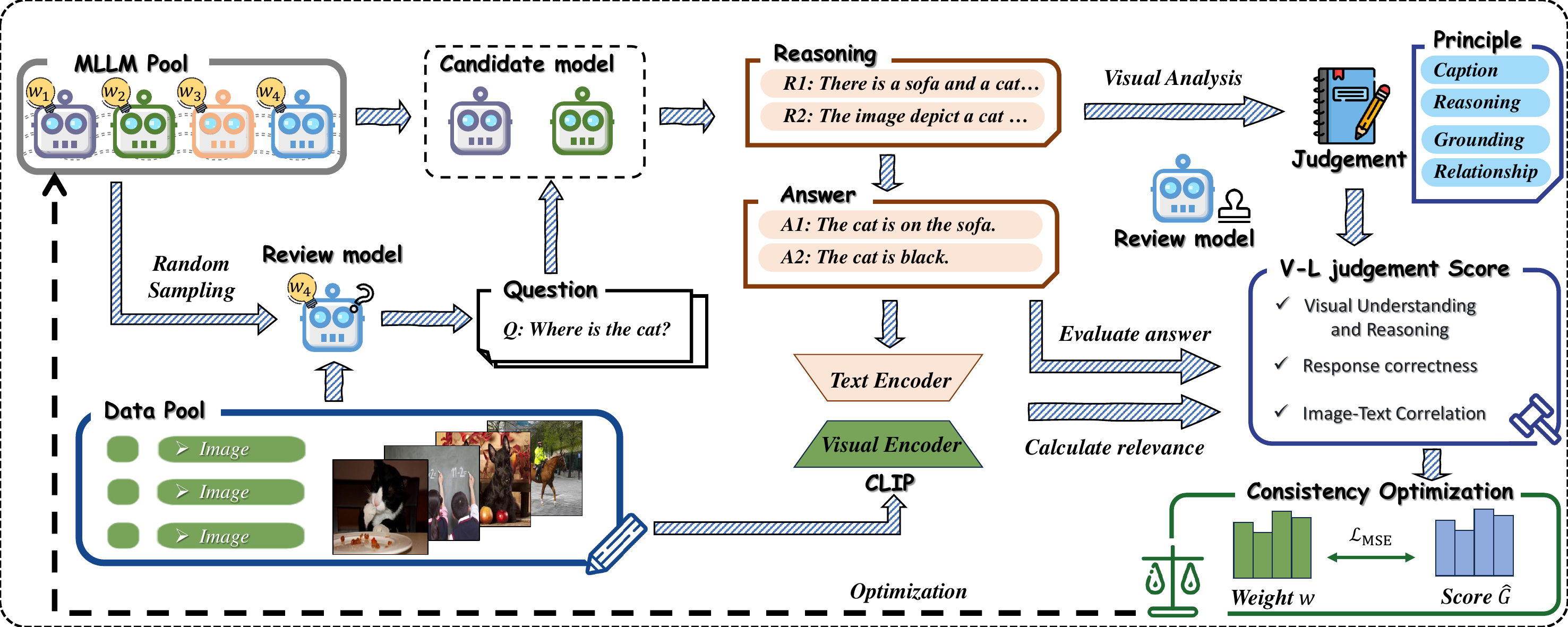}
    \caption{The UPME framework consists of three main components: \((i)\) Peer Review Mechanism, where two candidate models and one review model are randomly selected from the MLLM pool. The review model generates questions based on a selected image, and candidate models provide responses. \((ii)\) Vision-Language Judgment Scoring System, which evaluates answers based on textual correctness, visual understanding and reasoning, and image-text correlation. \((iii)\) Dynamic Weight Optimization, ensuring consistency between confidence weights and estimated scores through iterative optimization cycles.}
  \label{fig:short}
\end{figure*}

\subsection{MLLM-as-a-Judge}

To reduce the human annotation workload, LLM-as-a-Judge~\cite{zheng2023judging} is proposed to enable models to perform mutual evaluation. This approach primarily involves using LLMs to assess responses in two ways: comparing pairs of answers to determine superiority \citep{zheng2023judging,kasner2024traditionalbenchmarksanalyzingbehaviors,liu2023alignbenchbenchmarkingchinesealignment}, or directly scoring individual answers based on specific criteria \cite{gao2024best, liu2023alignbenchbenchmarkingchinesealignment}.
This strategy is extended to MLLM-as-a-Judge~\cite{chen2024mllm} to assess the performance of MLLMs across multiple evaluation tasks. MJ-Bench~\cite{chen2024mj} evaluates the effectiveness of multimodal reward models in text-to-image generation. However, these methods face two main biases~\cite{ye2024justice,raina2024llm}: the verbose bias and self-preference bias, which means MLLMs prefer the answers from themselves or with longer text length. Human preferences have also not been effectively incorporated into these automated evaluation systems~\cite{ning2024peer}.

\subsection{Human Preference-based Evaluation}

Predefined image and QA-pair benchmarks for MLLMs may encounter issues like data leakage and targeted optimization, which can prevent an accurate reflection of the true capabilities of these models \cite{deng2023benchmark,xu2024benchmarking}. Therefore, evaluation methods based on human preferences offer a better assessment of whether MLLMs meet human needs \cite{moell2024evaluating}. Chatbot Arena~\cite{chiang2024chatbot} provides a public platform specifically for evaluating LLMs based on human preference. This platform uses a pairwise comparison approach, where users interact with two anonymous models and select the one that performs better, generating a dynamic leaderboard via the Elo rating system. Although open-ended user evaluations can more realistically reflect LLM performance in real-world scenarios~\cite{wang2024userbenchmark}, the high economic cost of collecting large volumes of high-quality user feedback, combined with the long wait time for obtaining such feedback, limits the widespread adoption of this approach.

\section{Methodology}
\label{method}

Assigning higher weights to better-performing models leads to a significant improvement in the accuracy of the entire evaluation framework. This conclusion is supported by preliminary experiments detailed in~\autoref{differentW} and confirmed by recent research~\cite{ning2024peer, chu2024pre, tan2024judgebench, stephan2024calculation}. Therefore, we focus on characterizing the relationship between model performance and its judging capability through consistency optimization.

\subsection{Peer review formulation}

\paragraph{Problem definition.} The UPME framework facilitates mutual evaluation among MLLMs via a peer review mechanism, aiming to produce evaluation scores that closely reflect human preferences. Given an unsupervised image dataset $\mathcal{I} = \{I_i\}_{i=1}^n$ of $n$ images (with no human annotated questions and answers) and a model pool $\mathcal{M}=\{M_j\}_{j=1}^m$ consisting of $m$ MLLMs (both open-source and closed-source), we can obtain an UPME-learned score list $\hat{G}$:

\begin{equation}
\label{eq:upme_scores}
    \hat{G} := [\hat{G}_{M_1}, \hat{G}_{M_2}, \hat{G}_{M_3}, \dots, \hat{G}_{M_m}].
\end{equation}

On the other hand, given $G_{M_1}$ represents the score of model $M_1$ on the human-annotated benchmark, we define the human-annotated score list $G$ as:

\begin{equation}
\label{eq:human_scores}
    G := [G_{M_1}, G_{M_2}, G_{M_3}, \dots, G_{M_m}].
\end{equation}

Our objective is to optimize the UPME-learned scores $\hat{G}$ by maximizing their similarity to the human-annotated score list $G$, i.e., $\text{Sim}(G, \hat{G}) \rightarrow 1$, ensuring that UPME-learned scores align closely with human benchmark standards.

\paragraph{Overview of UPME.}
\autoref{fig:short} shows an overview of our UPME framework. During each review iteration, we randomly select two different candidate models \( M_j \) and \( M_k \) from the pool \(\mathcal{M}\) for comparison, along with another model \( M_r \) to serve as the reviewer. \( M_r \) generates a question \( Q_{i}^{r} = M_r(I_i) \) for each image \( I_i \) and then prompts the candidate models \( M_j \) and \( M_k \) to provide responses \( A_i^{j,r} \) and \( A_i^{k,r} \), respectively. The answers can be written as:
\begin{equation}
A_i^{j,r}=M_{j}\left(I_i, Q_{i}^{r}\right), \quad A_i^{k,r}=M_{k}\left(I_i, Q_{i}^{r}\right).
\end{equation}

Based on the vision-language judgment scoring System \( S_{VL} \) (detailed in~\autoref{3.2system}), \( M_r \) then evaluates these responses, and compares the responses of \( M_j \) and \( M_k \) to produce a review score \(\text{Review}_i^{j,k,r}\) as:

% \vspace{-3pt}
\begin{equation}
\label{eq: judge}
\text{Review}_i^{j,k,r} =  S_{VL}( A_i^{j,r},  A_i^{k,r}, Q_{i}^{r}, I_i \mid  M_r).
\end{equation}

By assigning each model an initiated confidence weight \( w \), we can calculate the learned score \(\hat{G}_{M_j}\) of each MLLM:
\begin{align}
\hat{G}_{M_j} = \sum_{i}\sum_{k \neq j}\sum_{r \neq k, r \neq j} \text{Review}_i^{j,k,r} \times w_r,
\label{eq:G}
\end{align}
then we can subsequently obtain the UPME-learned score list $\hat{G}$, as illustrated in~\autoref{eq:upme_scores}.

\subsection{Vision-Language Judgment Scoring System}
\label{3.2system}
In this subsection, we present our vision-language judgment scoring system, denoted as \( S_{\text{VL}} \) in~\autoref{eq9}. Our evaluation is structured around three core criteria: Response Correctness, Visual Understanding and Reasoning, and Image-Text Correlation. The first two aspects are evaluated by employing \( M_r \), which reviews responses based on prompts outlined in ~\autoref{sec:prompt_template}. The Image-Text Correlation is measured using CLIP scores.

\vspace{-5pt}

\paragraph{Response Correctness.} The primary criterion is the correctness of the candidate models' responses to questions posed by the review model. Existing studies indicate that paired comparative evaluation yields better accuracy than isolated assessments \cite{liusie2023zero, liu2024aligning}, so we adopt a pairwise scoring approach. Based on the review model \( M_r \)'s evaluation \( \text{Judge}_i^{r} \) for candidates \( M_j \) and \( M_k \), we calculate the model’s score for Response Accuracy \( S_{\text{Correct}}(j) \) as follows:
\begin{equation}
S_{\text{Correct}}(j) = \begin{cases}
1 & \text{if } \text{Judge}_{\text{Correct}, j} > \text{Judge}_{\text{Correct}, k} \\
0.5 & \text{if } \text{Judge}_{\text{Correct}, j} = \text{Judge}_{\text{Correct}, k} \\
0 & \text{if } \text{Judge}_{\text{Correct}, j} < \text{Judge}_{\text{Correct}, k},
\end{cases}
\label{eq6}
\end{equation}
where \( \text{Judge}_{\text{Correct}} \) represents the review model’s assessment of response correctness. If \( \text{Judge}_{\text{Correct}, j} > \text{Judge}_{\text{Correct}, k} \), it indicates that model \( M_j \) is judged superior in response accuracy to \( M_k \), scoring 1 point; a tie yields 0.5 points, and a lower score results in 0 points.

\vspace{-5pt}

\paragraph{Visual Understanding and Reasoning.} Inspired by the significant CoT ability demonstrated in the o1 model \cite{openai2024o1} through adversarial training, we leverage $M_r$ to evaluate the interpretative and analytical capabilities of the model with respect to visual content.

\( S_{\text{Visual}}(j) \) is computed using the function \(\Gamma\), which evaluates the response across four dimensions: caption, reasoning, grounding, and relationship, as follows:

\vspace{-3pt}

\begin{equation}
S_{\text{Visual}}(j) = \Gamma(\text{cap.}, \text{rea.}, \text{gro.}, \text{obj.} \mid A_j, M_r), 
\end{equation}
% captioning, reasoning, grounding, and objecting
where captioning evaluates the ability to generate precise descriptions, reasoning measures logical consistency, grounding assesses accurate object localization, and relationship captures the connections between subjects. Within this structure, the review model \( M_r \) provides a combined score via \(\Gamma\), yielding \( S_{\text{Visual}}(j) \). \( S_{\text{Visual}}(j) \) reflects a model’s capability to go beyond basic VQA, integrating both description and reasoning to demonstrate a deeper level of visual comprehension.

\vspace{-5pt}

\paragraph{Image-Text Correlation.} To capture the alignment between image content and textual responses, we introduce the CLIP \cite{radford2021learning} score:
\begin{equation}
S_{\text{Clip}}(j) = \text{CLIP}(I_i, A_i^{j,r}),
\end{equation}
where \( S_{\text{Clip}}(j) \) measures the correlation between image \( I_i \) and response \( A_i^{j,r} \) from model \( M_j \). The details are described in ~\autoref{detailsofClip}.

Combining the above components, we compute the final vision-language judgment score as a weighted sum of the three criteria: Response Accuracy, Visual Understanding and Reasoning, and Image-Text Correlation:
\begin{equation}
S_{\text{VL}} = \gamma_1 S_{\text{Correct}} + \gamma_2 S_{\text{Visual}} + \gamma_3 S_{\text{Clip}},
\label{eq9}
\end{equation}
where \( \gamma_1 \), \( \gamma_2 \), and \( \gamma_3 \) are weights reflecting the relative importance of each component in the overall evaluation.

\begin{table*}[ht]
    \centering
    \small
    \resizebox{0.87\textwidth}{!}{ % 设置宽度为 textwidth 的 1 倍
        \setlength{\tabcolsep}{3pt}
        \setlength{\tabcolsep}{6pt} % 增加列间距
        \renewcommand{\arraystretch}{1.1}

        \begin{tabular}{l|ccc|ccc}
            \toprule[1.5pt]
            \multicolumn{1}{c|}{\multirow{2}*{\textbf{Models}}} & \multicolumn{3}{c|}{\textbf{MMstar}} & \multicolumn{3}{c}{\textbf{ScienceQA}} \\
            
            & \textbf{Pearson ($\uparrow$)} & \textbf{Spearman ($\uparrow$)} & \textbf{Per.Ent. ($\downarrow$)} & \textbf{Pearson ($\uparrow$)} & \textbf{Spearman ($\uparrow$)} & \textbf{Per.Ent. ($\downarrow$)} \\
            \midrule
            LLama-3.2-11b-V & 0.314$^{\pm0.0757}$ & 0.550$^{\pm0.0577}$ & 0.983$^{\pm0.2310}$ & 0.160$^{\pm0.0430}$ & 0.225$^{\pm0.0957}$ & 1.099$^{\pm0.0000}$ \\
            Claude-3-haiku & 0.095$^{\pm0.1301}$ & 0.225$^{\pm0.1500}$ & 1.099$^{\pm0.0000}$ & -0.145$^{\pm0.0308}$ & -0.525$^{\pm0.2061}$ & 1.099$^{\pm0.0000}$ \\
            Claude-3.5-sonnet & 0.780$^{\pm0.0146}$ & 0.825$^{\pm0.0500}$ & 0.752$^{\pm0.2310}$ & 0.437$^{\pm0.0452}$ & 0.450$^{\pm0.1732}$ & 0.752$^{\pm0.2310}$ \\
            Gemini-1.5-pro & 0.864$^{\pm0.0068}$ & 0.850$^{\pm0.0577}$ & 0.637$^{\pm0.0000}$ & 0.4147$^{\pm0.0464}$ & \textbf{0.725}$^{\pm0.2062}$ & \textbf{0.434}$^{\pm0.5352}$ \\
            GPT-4o-mini & 0.668$^{\pm0.0073}$ & 0.725$^{\pm0.1258}$ & 0.868$^{\pm0.2886}$ & 0.3354$^{\pm0.0635}$ & 0.600$^{\pm0.0000}$ & 1.099$^{\pm0.0000}$ \\
            GPT-4o & \textbf{0.878}$^{\pm0.0038}$ & \textbf{0.875}$^{\pm0.0050}$ & \textbf{0.159}$^{\pm0.3183}$ & \textbf{0.617}$^{\pm0.0071}$ & 0.625$^{\pm0.1258}$ & 0.637$^{\pm0.0000}$ \\
            \midrule
            \multicolumn{1}{c}{\textbf{Methods}} & & & & & & \\
            \midrule
            Peer Review & 0.725$^{\pm0.0044}$ & 0.771$^{\pm0.1616}$ & 1.040$^{\pm0.2830}$ & 0.463$^{\pm0.0193}$ & 0.686$^{\pm0.1777}$ & 1.040$^{\pm0.2830}$ \\
            Majority Vote \cite{surowiecki2005wisdom} & 0.757$^{\pm0.0013}$ & 0.757$^{\pm0.0857}$ & 1.299$^{\pm0.1733}$ & 0.509$^{\pm0.0181}$ & 0.524$^{\pm0.0660}$ & 1.040$^{\pm0.0000}$ \\
            Rating Vote \cite{allahbakhsh2012rating} & 0.795$^{\pm0.0013}$ & 0.743$^{\pm0.2309}$ & 0.628$^{\pm0.0755}$ & 0.623$^{\pm0.0084}$ & 0.629$^{\pm0.1895}$ & 0.920$^{\pm0.2387}$ \\
            PRD \cite{li2023prd} & 0.838$^{\pm0.0027}$ & 0.864$^{\pm0.0317}$ & 0.427$^{\pm0.0087}$ & 0.636$^{\pm0.0042}$ & 0.694$^{\pm0.0734}$ & 0.746$^{\pm0.0016}$\\
            UPME & \underline{\textbf{0.944$^{\pm0.0011}$}} & \underline{\textbf{0.972$^{\pm0.0330}$}} & \underline{\textbf{0.141$^{\pm0.2812}$}} & \underline{\textbf{0.814$^{\pm0.0024}$}} & \underline{\textbf{0.886$^{\pm0.0286}$}} & \underline{\textbf{0.422$^{\pm0.2812}$}} \\
            \bottomrule[1.5pt]
        \end{tabular}
    }
    \caption{Main results of different models and methods on MMstar and ScienceQA datasets, evaluated by Pearson, Spearman, and Permutation Entropy metrics. Since the model cannot judge itself, scores from the other five models involved in the computations are used for evaluation. Optimal results are highlighted in bold and underlined. Standard deviation (std) is calculated based on four runs.}
    \label{tab:main}
\end{table*}

\subsection{Dynamic Weight Optimization}
In each iteration, we use Mean Squared Error (MSE) as the loss function to update the weights  \( w \), enhancing the consistency between the weights  \( w \) and the estimated scores \( \hat{G} \):
\begin{equation}
    \mathcal{L}_\text{MSE}(\hat{G}, w) = \frac{1}{m} \sum_{j=1}^{m} \left(\hat{G}_{M_j} - w_{M_j}\right)^2.
\label{eq:optim}
\end{equation}

The entire optimization process is dynamic. After each round of weight updates, the weights and scores continue to change. The loss function computes new scores based on the updated weights, which are then iteratively used to optimize the weights further, forming a continuous iterative optimization loop, as detailed in~\autoref{alg: UPME}.

\section{Experiments}

\subsection{Experimental Setup}

\noindent \textbf{Datasets.} To validate the effectiveness of our framework, we conduct experiments on two distinctly styled datasets: ScienceQA~\cite{lu2022learn} and MMStar~\cite{chen2024we}. ScienceQA underscores the importance of CoT in scientific reasoning tasks, comprising approximately 21,208 multimodal multiple-choice questions covering various science topics. MMStar focuses on evaluating the multimodal capabilities of MLLMs, and consists of 1,500 meticulously curated samples including 6 core capabilities and 18 detailed axes. Each sample exhibits visual dependency and minimal data leakage. 

\noindent \textbf{Candidate MLLMs.}
Our MLLM pool includes 5 closed-source models and 1 open-source model. The open-source model is Llama-3.2-11b-vision-instruct~\cite{meta2024llama3.2_11b}, which adapts the Llama-3 architecture to the multimodal domain through visual instruction tuning~\cite{liu2024visual}. The closed-source models consist of GPT-4o~\cite{openai_gpt4o_2024}, GPT-4o-mini~\cite{openai2024gpt4omini}, Claude-3.5-Sonnet~\cite{anthropic2024claude35}, Claude-3-Haiku~\cite{anthropic2024claude3haiku}, and Gemini-1.5-Pro~\cite{team2023gemini}. These commercial models employ enhanced human alignment strategies to optimize performance and user experience across various applications.

\noindent \textbf{Metrics.}
For all experiments, we employ three popular metrics to evaluate the setups mentioned above and our UPME method: the Pearson Correlation Coefficient~\cite{cohen2009pearson} to evaluate the similarity of score lists, Spearman's Rank Correlation Coefficient~\cite{zar2005spearman} to measure the correlation of ranking lists, and Permutation Entropy~\cite{bandt2002permutation} to assess the complexity and unpredictability of the scoring distributions. All experiments are performed for 4 runs with different seeds (\(seed = 1,2,3,4\) ) and record the average results.

\noindent \textbf{Comparison Methods.} Since we are the first to investigate unsupervised peer review in MLLM evaluations, we choose the original peer review mechanism and two popular classical methods, Majority Vote \cite{surowiecki2005wisdom} and Rating Vote \cite{allahbakhsh2012rating}, for our comparative analysis. We also adopt a semi-supervised method, PRD \cite{li2023prd}, and conduct experiments using individual models as judges to further bolster our comparisons.

\subsection{Experimental Results}

\subsubsection{Comparison Results}

\begin{figure}
    \centering
    \includegraphics[width=0.97\linewidth]{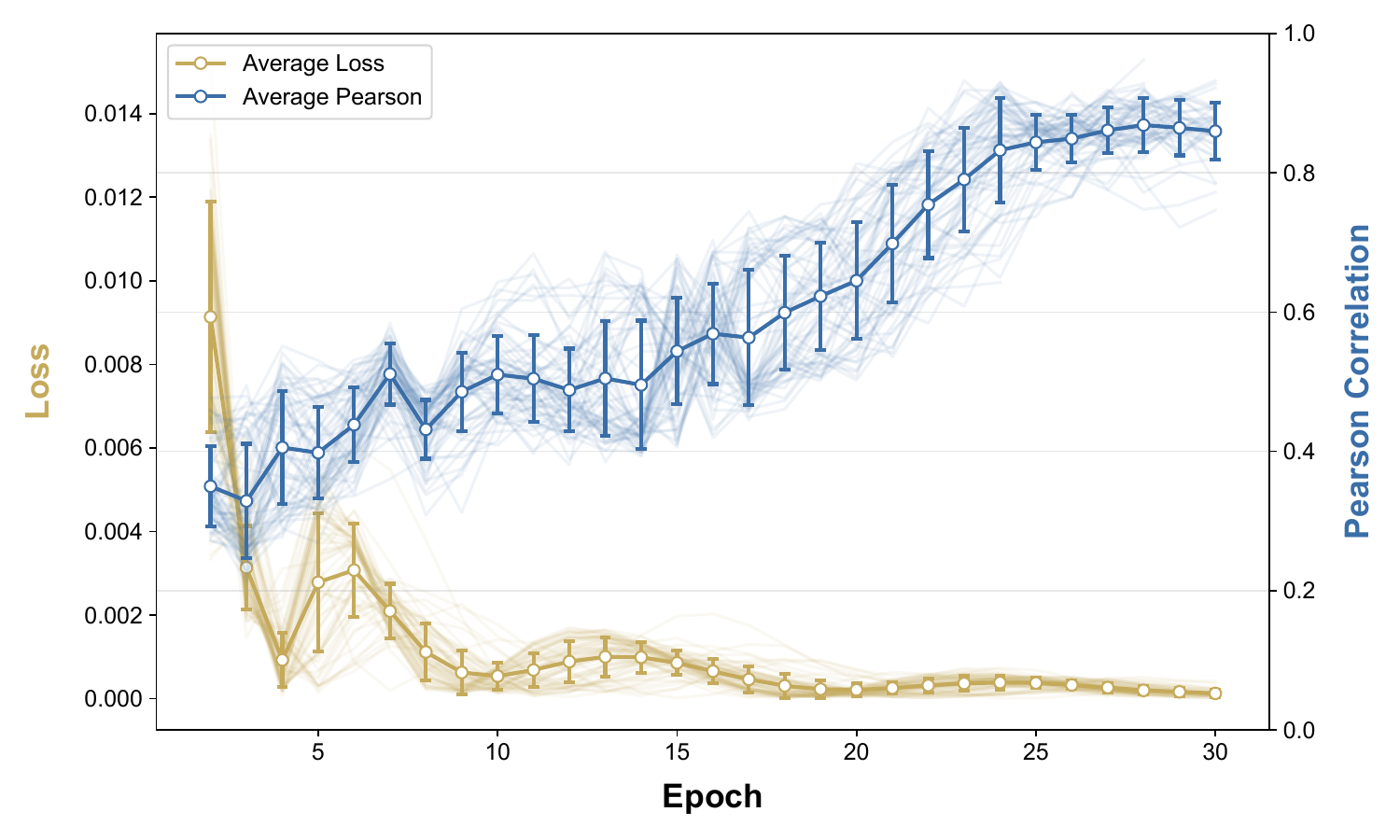}
    \vspace{-5pt}
    \caption{Convergence experiments.}
    \label{fig:convergence_experiment}
    \vspace{-5pt}
\end{figure}

The experimental results in \autoref{tab:main} indicate that among the candidate MLLMs, GPT-4o exhibits the highest review capability on the MMstar dataset, achieving Pearson and Spearman correlation coefficients of 0.878 and 0.875, respectively. In contrast, the open-source LLama-3.2 model demonstrates weaker performance. Although Claude-3.5-Sonnet and Gemini-1.5-Pro claim to offer performance comparable to GPT-4o, they still show limitations when evaluating other models, falling noticeably short of human-level performance. The original peer review method yields Pearson and Spearman coefficients lower than that of GPT-4o, indicating that the biases of weaker models influence the Peer Review approach. Our UPME method achieves a Pearson correlation of 0.944 and a Spearman correlation of 0.972 on the MMstar dataset, demonstrating a high degree of similarity with human annotations and validating the effectiveness of our optimizations in visual QA comprehension and consistency evaluation.

On the ScienceQA dataset, even the best-performing GPT-4o achieves Pearson and Spearman correlations of 0.617 and 0.625, reflecting the difficulty of the task. The original peer review method shows more modest results, with Pearson and Spearman of 0.463 and 0.686, respectively. Following a series of optimizations and alignments, UPME achieves Pearson and Spearman correlations of 0.814 and 0.886, showcasing its robustness and superior performance in complex datasets. While the performance on ScienceQA may not be as high, it is important to note that traditional datasets like ScienceQA might not fully align with the evaluation needs of current MLLMs because they may lack visual dependency or sufficient discrimination~\cite{yu2023mm, chen2024we}. UPME achieves higher Pearson and Spearman correlations, demonstrating its robustness and superior performance.

Moreover, as shown in~\autoref{fig:convergence_experiment}, the loss consistently decreases, and the Pearson similarity metric increases over epochs in 64 different initial settings, which demonstrates the convergence of $w$ and $\hat{G}$. Under the current experimental setup (25 images and 1500 judgments), the framework reliably converges within the 30-epoch limit.

\begin{table}[t]
    \centering
    \setlength\tabcolsep{1.0mm}
    \renewcommand\arraystretch{1.1}
    \def \mysp {\hspace{7pt}}
    {\small 
    \resizebox{\linewidth}{!}{
    \begin{tabular}{cccccccc}
    \toprule[1.5pt]
    & & & & \multicolumn{2}{c}{MMstar} & \multicolumn{2}{c}{ScienceQA} \\
    \toprule
         \makecell{Method} & Cor. & Vis. & Cli. & \textbf{Pearson} & \textbf{Spearman} & \textbf{Pearson} & \textbf{Spearman} \\
    \hline
        \makecell{Peer \\ Review} & & & & 0.727 & 0.714 & 0.457 & 0.600 \\
        \hline
        \makecell{Score \\ Optimization}  & \checkmark & & & 0.854 & 0.771 & 0.713 & 0.657 \\        
        \hline
        \multirow{3}{*}{{\makecell{Visual \\ Alignment}}}  &   & \checkmark & & 0.873 & 0.886 & 0.701 & 0.771 \\
        &  &  & \checkmark & 0.785 & 0.829 & 0.548 & 0.771\\
        &  & \checkmark & \checkmark & 0.903 & 0.943 & 0.775 & 0.886 \\
    \hline
    $\textbf{UPME}$ & \checkmark & \checkmark & \checkmark & \textbf{0.944} & \textbf{0.972} & \textbf{0.814} & \textbf{0.886} \\
    \bottomrule[1.5pt]
    \end{tabular}}
    }
    \caption{Ablation study with Pearson and Spearman similarity metrics for each dataset. Optimal results are highlighted in bold and underlined. `Cor.' stands for correctness, `Vis.' stands for visual understanding, and `Cli.' stands for Clip.}
    \label{tab:ablation}
    \vspace{-5pt}
\end{table}

\subsubsection{Ablation Study}

In the ablation study presented in \autoref{tab:ablation}, we evaluated the impact of progressively introducing different components on model performance. The Score Optimization phase showed consistent improvements in Pearson and Spearman correlation coefficients as modules were added. For instance, using only the correctness module resulted in Pearson scores of 0.854 for MMstar and 0.713 for ScienceQA. Introducing the visual understanding module alone yielded scores of 0.873 and 0.701, respectively. Adding the Clip component alone gave scores of 0.785 for MMstar and 0.548 for ScienceQA. Combining visual understanding and Clip modules further improved performance to 0.903 and 0.775. Ultimately, integrating all components within the UPME framework resulted in the highest performance, with Pearson and Spearman correlations of 0.944 and 0.972 on MMstar and 0.814 and 0.886 on ScienceQA. These results confirm that optimizing the scoring mechanism and visual alignment significantly boosts evaluation accuracy and robustness across different datasets.

\begin{figure}[t]
  \centering
   \includegraphics[width=\linewidth]{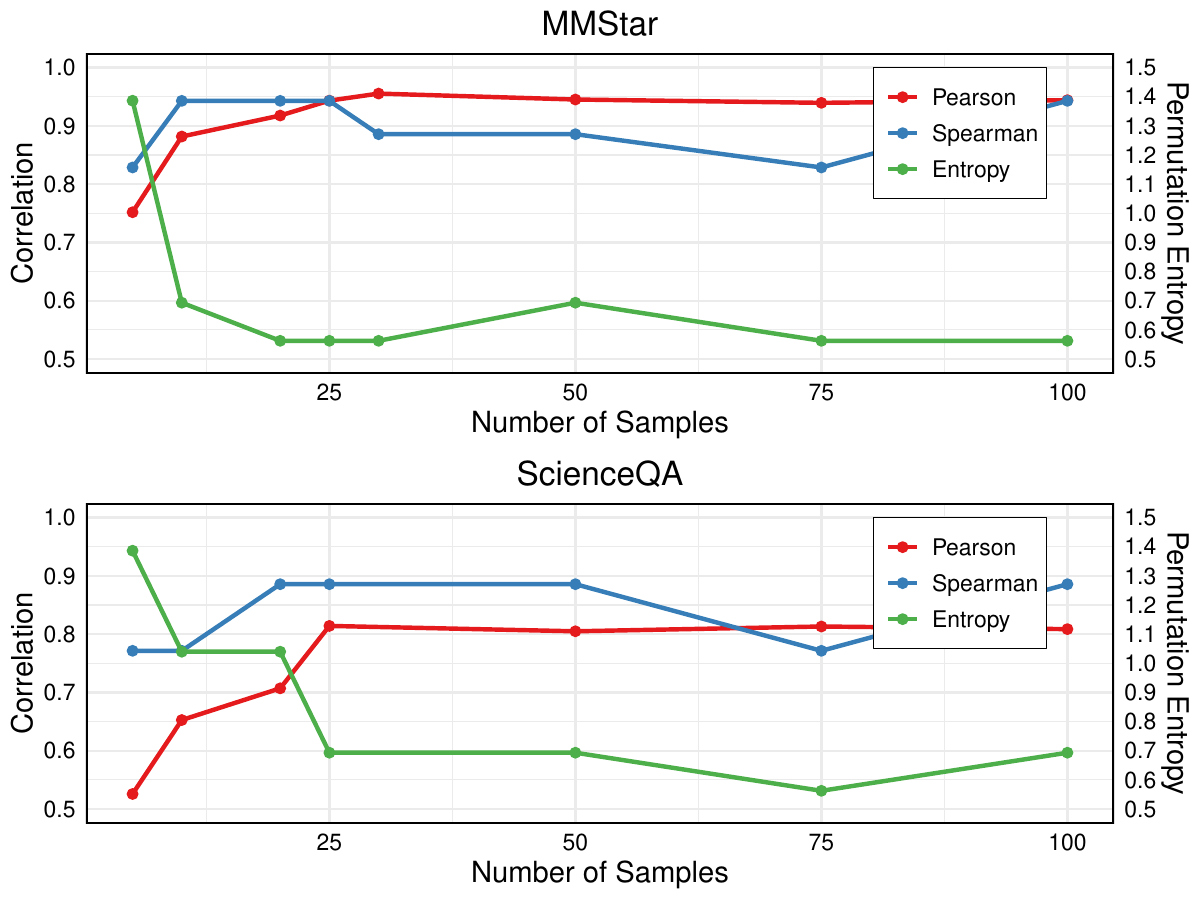}

   \caption{The performance of UPME in different sample size.}
   \label{fig:sample_size}
   \vspace{-5pt}
\end{figure}

\subsubsection{Impact of Sample Size}
In \autoref{fig:sample_size}, we analyze the impact of sample size on the alignment with human-annotated QA benchmarks, measured through Pearson and Spearman correlation coefficients, as well as Permutation Entropy. As the sample size increases from 5 to 25, there is a clear improvement in both correlation metrics and a decline in Permutation Entropy, indicating that the generated score lists and rank lists show greater alignment with human evaluations.
Beyond a sample size of 25, the metrics stabilize, showing minimal changes with further increases from 25 to 100 in sample size. This stabilization suggests that the consistency of the generated outputs relative to human judgments has reached a satisfactory level, making 25 samples a reliable point for evaluation. Therefore, we primarily selected a sample size of 25 for our experiments, as it provides an effective balance between alignment accuracy and data efficiency.

\section{Human Preference Alignment Analysis}
\label{humanalign}

The aforementioned experiments primarily reflect the consistency between the UPME framework and the predefined QA-based benchmarks. To further explore the alignment between UPME and human preferences, as well as to examine the two types of biases inherent in MLLM-as-Judge approaches, we selected a subset of responses for human evaluation and analysis of UPME scoring results. The detailed experimental design is provided in~\autoref{humanannotation}.

\subsection{Alignment with Human Preference}
\renewcommand\arraystretch{1.2}
\begin{table}[ht]
\centering

\resizebox{\linewidth}{!}{%
\begin{tabular}{cccc}
\toprule[1.5pt]
\textbf{Dataset}     & \textbf{Method}         & \textbf{Agreement (\%)} & \textbf{Consistency (\%)} \\
\hline
{\multirow{2}*{\textbf{MMstar}}}   & Peer Review             & 71.1         & 67.5                \\
                     & UPME         & 95.9                          & 89.8                            \\
\hline
{\multirow{2}*{\textbf{ScienceQA}}}   & Peer Review             & 68.2           & 61.8                \\
                     & UPME         & 87.4                          & 82.6                            \\
\bottomrule[1.5pt]

\end{tabular}
}
\caption{Human Agreement (\%) and Human Consistency (\%) Rates on Various Datasets.}
\label{tab:humanalignment}
\vspace{-5pt}
\end{table}

To validate whether our proposed UPME method can align with human preferences without manual labeling, we conducted experiments on the MMstar and ScienceQA datasets. The experimental results on \autoref{tab:humanalignment} indicate that the baseline method exhibited relatively low human agreement and model consistency rates, suggesting that the Peer Review mechanism under an unsupervised setting without weight optimization struggles to align with human preferences. In contrast, UPME demonstrated significant improvements by incorporating  Correctness, Visual Understanding, and Clip Correlation. On the MMStar, UPME achieved an agreement rate of 95.9\% and a consistency rate of 89.8\%, showing that the optimized scoring criteria significantly enhance the accuracy of evaluation outputs and alignment with human preferences. By capturing key multimodal understanding metrics without relying on manual labeling, UPME effectively achieved high consistency with human annotations, highlighting its substantial advantages in improving response consistency and accuracy under an unsupervised framework.

\begin{figure}[t]
  \centering
   \includegraphics[width=\linewidth]{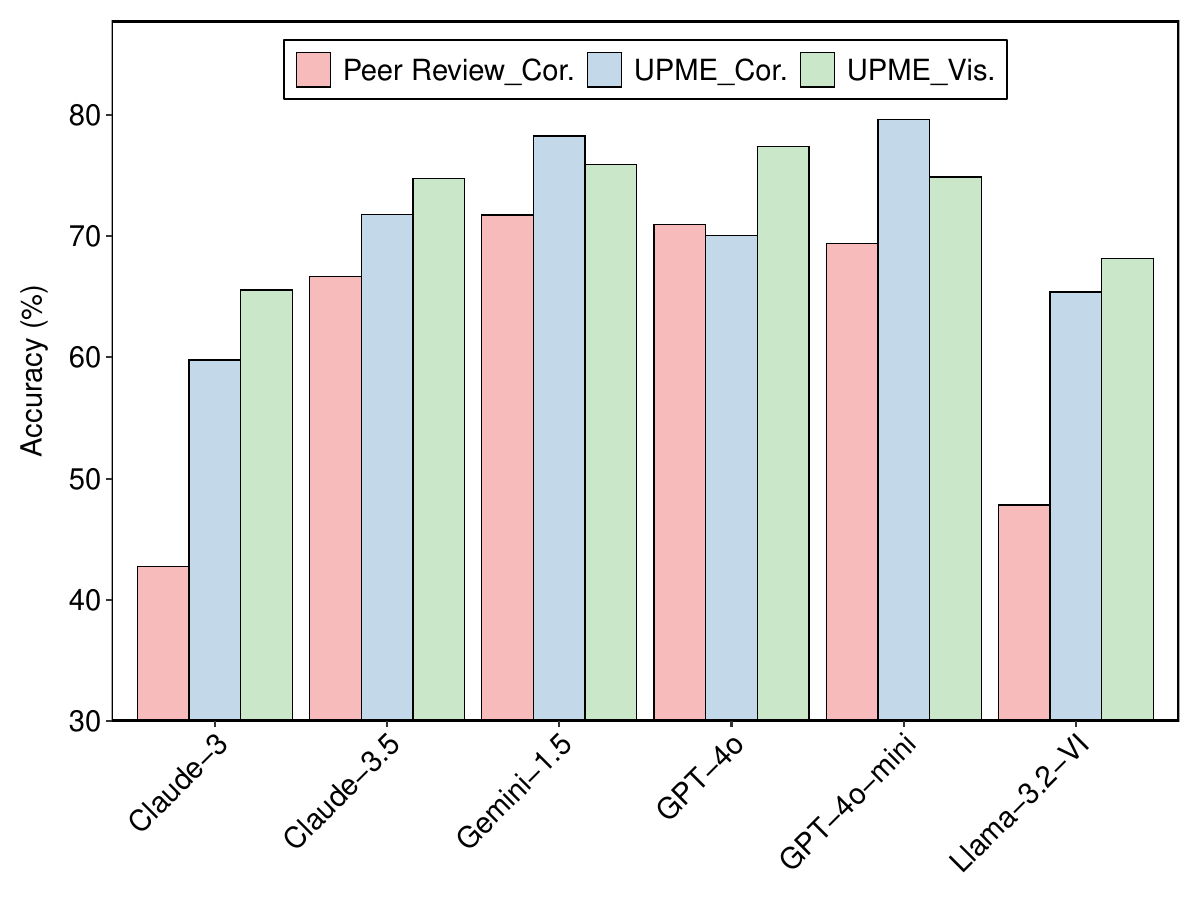}

   \caption{Model accuracy comparison in peer review framework w/ and w/o UPME, where \textit{Peer Review\_Cor.} represents the correctness of the original peer review, and \textit{UPME\_Cor.} and \textit{UPME\_Vis.} correspond to the two judgment dimensions of response correctness and visual understanding, introduced in \autoref{3.2system}.}
   \label{fig:UPME_acc_model}
\end{figure}

\subsection{Performance w/ and w/o UPME}

To assess the improvement in model judgment accuracy within the Peer Review mechanism and the UPME framework, we conducted an experiment, the results are displayed in \autoref{fig:UPME_acc_model}. The calculation of accuracy is based on $ACC = \frac{\text{judge}_\text{correct}}{\text{judge}_\text{all}}$. Experimental results show that the UPME significantly improves the accuracy of models in both correctness and visual understanding compared to the original correctness judgment. Specifically, the average accuracy in correctness evaluation increased from 61.56\% in the baseline to 74.48\% with UPME, an improvement of 12.92\%. This indicates a significant advantage of UPME in assessing the correctness of model responses. Furthermore, in the evaluation of visual understanding, UPME achieved an average accuracy of 73.93\%, which is also notably higher than the baseline. This improvement is primarily attributed to UPME's unsupervised evaluation framework, which employs a multi-model peer review mechanism that combines correctness and visual understanding scores to achieve a more efficient and accurate assessment.

\begin{figure}[t]
  \centering
  \begin{minipage}{0.49\linewidth}
    \centering
    \includegraphics[width=\linewidth]{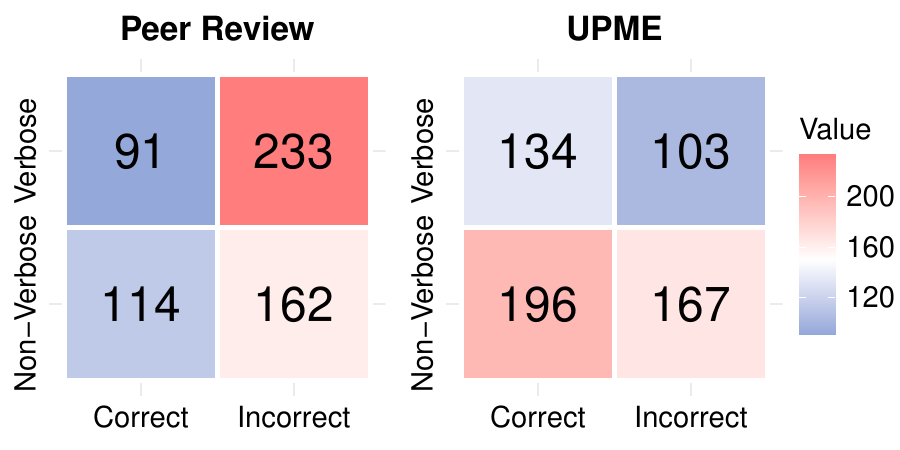}
    \caption*{(a) Verbosity Bias}
  \end{minipage}%
  \hfill
  \begin{minipage}{0.49\linewidth}
    \centering
    \includegraphics[width=\linewidth]{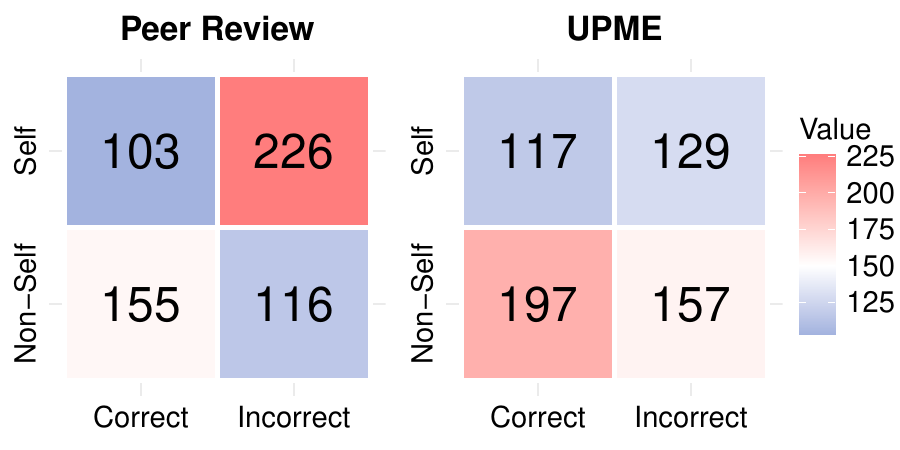}
    \caption*{(b) Self Preference}
  \end{minipage}
  
  \vspace{1em}
  
  \centering
  \small
  \resizebox{\linewidth}{!}{
    \begin{tabular}{cccccc}
      \toprule[1.5pt]
      \textbf{Bias Type} & \textbf{Method} & \textbf{Chi-Square} & \textbf{p-value} & \textbf{Phi Coefficient} \\
      \midrule
      \multirow{2}{*}{{\textbf{\makecell{Verbosity \\ Bias}}}} & Peer review & 10.996 & 0.00091 & 0.135 \\

      & UPME         & 0.280  & 0.59696 & 0.022 \\
      \midrule
      \multirow{2}{*}{{\textbf{\makecell{Self \\ Preference}}}} & Peer review & 39.584 & 3.142e-10 & 0.257 \\
      & UPME         & 3.489  & 0.0618    & 0.076 \\
      \bottomrule[1.5pt]
    \end{tabular}}
  
  \caption{Heatmap and Table for Peer Review and UPME on Self Preference and Verbosity Bias.}
  \label{fig:combined_heatmap_table}
\end{figure}

\begin{figure*}[!t]
    \centering
    \includegraphics[width=\textwidth]{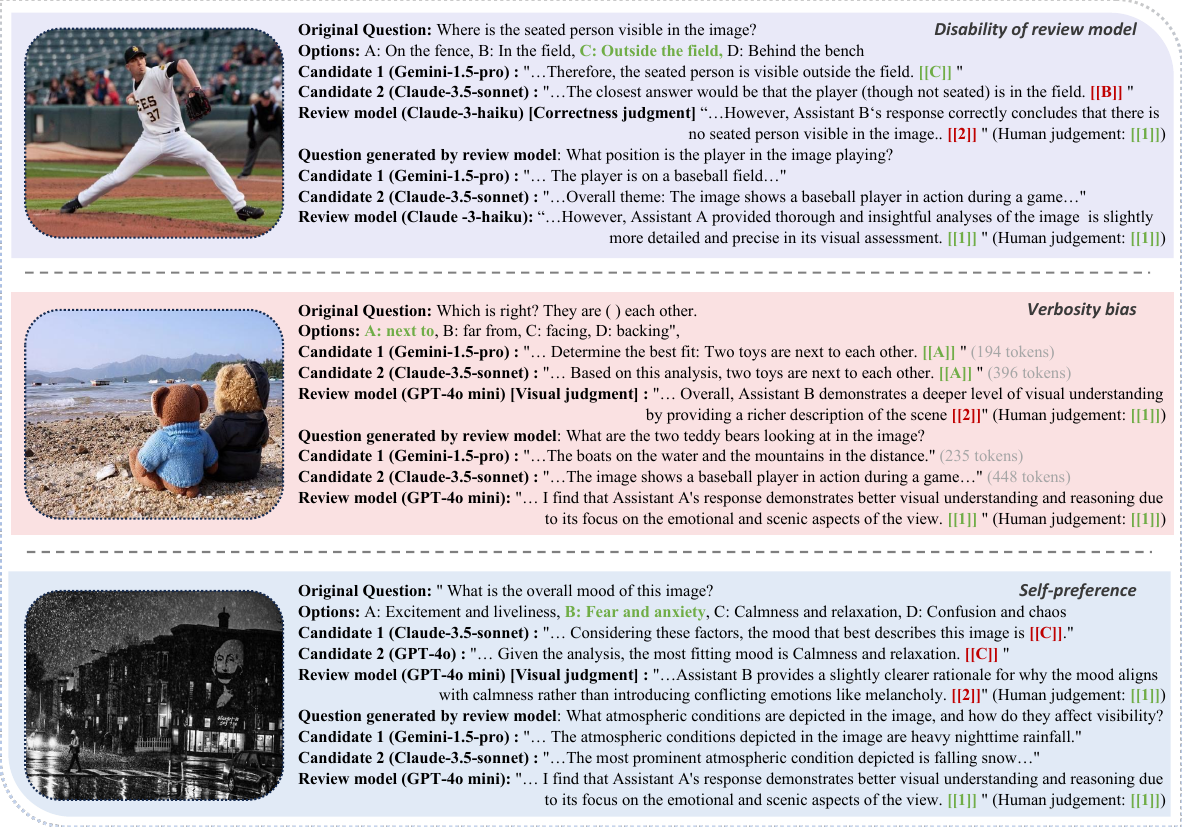}
    \caption{Case study illustration of UPME. We provide the original human-designed questions and the UPME-generated questions, along with the answer analysis. The upper case presents the \textit{Disability of review model}, where the review model can not answer the original question itself. The middle case demonstrates cases exhibiting \textit{verbosity bias}. The bottom case shows \textit{self-preference bias}.}

    \label{fig:case}
\end{figure*}

\subsection{Mitigating the Bias of MLLM-as-a-Judge}

\paragraph{Mitigation of Verbosity bias.} As shown in ~\autoref{fig:combined_heatmap_table}, the verbosity-preference bias is significantly alleviated by our method. The original Peer Review approach had a p-value of 0.00091 ($< 0.05$), with a Chi-square value of 10.996 and a Phi coefficient of 0.135, suggesting a significant preference for verbose responses during evaluation. In contrast, the UPME framework yielded a p-value of 0.59696 ($> 0.05$), with the Chi-square value reduced to 0.280 and the Phi coefficient decreased to 0.022, indicating that it does not exhibit a significant bias toward selecting verbose responses. Additionally, the number of incorrect selections of verbose responses dropped from 233 to 103 in the UPME framework. Thus, the UPME approach effectively mitigates the bias toward verbose responses in models, enhancing evaluation diversity and alignment with human preferences in an unsupervised setting.

\vspace{-10pt}
\paragraph{Mitigation of Self-preference.} Self-preference bias is also significantly alleviated, as shown in ~\autoref{fig:combined_heatmap_table}. In the original Peer Review method, the Chi-square value was 39.584, the Phi coefficient was 0.257, and the p-value was 3.142e-10, indicating a highly significant bias. In contrast, the UPME framework reduced the Chi-square value to 3.489, the Phi coefficient to 0.076, and the p-value to 0.0618, approaching insignificance. The incorrect selections due to self-preference dropped from 226 to 129, further demonstrating our effectiveness in mitigating this bias.

\subsection{Case Study}

We selected three cases to demonstrate how the UPME effectively addresses the core issues in traditional MLLM-as-a-judge as shown in \autoref{fig:case}. In the first case, the review model initially failed to identify the error in Candidate B's response because it was unable to answer the question itself. After applying UPME, the review model was able to generate a question within its capabilities and accurately determine that Candidate A's response showed a deeper understanding of the image, which aligned with human preferences, thus avoiding the misjudgment issue. In the second case, although Candidate B's response was longer, it did not provide additional useful information. Traditional evaluations might wrongly consider the longer answer as more informative, but with UPME, the review model, through precise text evaluation, clearly identified Candidate A's concise and clear answer as more valuable, thus avoiding verbosity bias. Finally, in the third case, UPME effectively avoided self-preference bias, enhancing fairness and objectivity.
\vspace{9pt}
\section{Conclusion}
This study proposes an unsupervised peer review-based framework, UPME, effectively reducing the annotation workload in traditional human-designed 
QA benchmarks. Our vision-language scoring system emphasizes a vision-oriented judgment, addressing issues such as verbosity and self-preference biases inherent in MLLM-as-judge evaluation methods. Further experiments validate that our evaluation results achieve higher alignment with human preferences, providing a promising and reliable approach.

\clearpage
\newpage

\section*{Acknowledgement}
This work was supported by Alibaba DAMO Academy through the Alibaba Innovative Research Program.

{
    \small
    \bibliographystyle{ieeenat_fullname}
    \bibliography{main}

\begin{thebibliography}{72}
\providecommand{\natexlab}[1]{#1}
\providecommand{\url}[1]{\texttt{#1}}
\expandafter\ifx\csname urlstyle\endcsname\relax
  \providecommand{\doi}[1]{doi: #1}\else
  \providecommand{\doi}{doi: \begingroup \urlstyle{rm}\Url}\fi

\bibitem[Achiam et~al.(2023)Achiam, Adler, Agarwal, Ahmad, Akkaya, Aleman, Almeida, Altenschmidt, Altman, et~al.]{achiam2023gpt}
Josh Achiam, Steven Adler, Sandhini Agarwal, Lama Ahmad, Ilge Akkaya, Florencia~Leoni Aleman, Diogo Almeida, Janko Altenschmidt, Sam Altman, et~al.
\newblock Gpt-4 technical report.
\newblock \emph{arXiv preprint arXiv:2303.08774}, 2023.

\bibitem[Allahbakhsh and Ignjatovic(2012)]{allahbakhsh2012rating}
Mohammad Allahbakhsh and Aleksandar Ignjatovic.
\newblock Rating through voting: An iterative method for robust rating.
\newblock \emph{arXiv preprint arXiv:1211.0390}, 2012.

\bibitem[Alpher(2002)]{Alpher02}
FirstName Alpher.
\newblock Frobnication.
\newblock \emph{IEEE TPAMI}, 12\penalty0 (1):\penalty0 234--778, 2002.

\bibitem[Alpher and Fotheringham-Smythe(2003)]{Alpher03}
FirstName Alpher and FirstName Fotheringham-Smythe.
\newblock Frobnication revisited.
\newblock \emph{Journal of Foo}, 13\penalty0 (1):\penalty0 234--778, 2003.

\bibitem[Alpher and Gamow(2005)]{Alpher05}
FirstName Alpher and FirstName Gamow.
\newblock Can a computer frobnicate?
\newblock In \emph{CVPR}, pages 234--778, 2005.

\bibitem[Alpher et~al.(2004)Alpher, Fotheringham-Smythe, and Gamow]{Alpher04}
FirstName Alpher, FirstName Fotheringham-Smythe, and FirstName Gamow.
\newblock Can a machine frobnicate?
\newblock \emph{Journal of Foo}, 14\penalty0 (1):\penalty0 234--778, 2004.

\bibitem[Anthropic(2024{\natexlab{a}})]{anthropic2024claude35}
Anthropic.
\newblock Claude 3.5: A sonnet.
\newblock \url{https://www.anthropic.com/news/claude-3-5-sonnet}, 2024{\natexlab{a}}.

\bibitem[Anthropic(2024{\natexlab{b}})]{anthropic2024claude3haiku}
Anthropic.
\newblock Claude 3 haiku.
\newblock \url{https://www.anthropic.com/news/claude-3-haiku}, 2024{\natexlab{b}}.

\bibitem[Anthropic(2024{\natexlab{c}})]{anthropic2024claudePCoperationDevelop}
Anthropic.
\newblock Developing a computer use model, 2024{\natexlab{c}}.

\bibitem[Anthropic(2024{\natexlab{d}})]{anthropic2024claudePCoperationnews}
Anthropic.
\newblock Introducing computer use, a new claude 3.5 sonnet, and claude 3.5 haiku, 2024{\natexlab{d}}.

\bibitem[Bandt and Pompe(2002)]{bandt2002permutation}
Christoph Bandt and Bernd Pompe.
\newblock Permutation entropy: a natural complexity measure for time series.
\newblock \emph{Physical review letters}, 88\penalty0 (17):\penalty0 174102, 2002.

\bibitem[Bao et~al.(2024)Bao, Huang, Wang, Ye, Wang, Chen, Elhoseiny, and Zhang]{bao2024autobenchv}
Han Bao, Yue Huang, Yanbo Wang, Jiayi Ye, Xiangqi Wang, Xiuying Chen, Mohamed Elhoseiny, and Xiangliang Zhang.
\newblock Autobench-v: Can large vision-language models benchmark themselves?
\newblock \emph{arXiv preprint arXiv:2410.21259}, 2024.

\bibitem[Becker et~al.(2023)Becker, Denny, Finnie-Ansley, Luxton-Reilly, Prather, and Santos]{becker2023programming}
Brett~A Becker, Paul Denny, James Finnie-Ansley, Andrew Luxton-Reilly, James Prather, and Eddie~Antonio Santos.
\newblock Programming is hard-or at least it used to be: Educational opportunities and challenges of ai code generation.
\newblock In \emph{Proceedings of the 54th ACM Technical Symposium on Computer Science Education V. 1}, pages 500--506, 2023.

\bibitem[Cai et~al.(2023)Cai, Mao, Wu, Wang, Liang, Ge, Wu, You, Song, Xia, et~al.]{cai2023low}
Yuzhe Cai, Shaoguang Mao, Wenshan Wu, Zehua Wang, Yaobo Liang, Tao Ge, Chenfei Wu, Wang You, Ting Song, Yan Xia, et~al.
\newblock Low-code llm: Visual programming over llms.
\newblock \emph{arXiv preprint arXiv:2304.08103}, 2, 2023.

\bibitem[Chen et~al.(2024{\natexlab{a}})Chen, Chen, Zhang, Liu, Wang, Zhou, Zhang, Zhou, Wan, and Sun]{chen2024mllm}
Dongping Chen, Ruoxi Chen, Shilin Zhang, Yinuo Liu, Yaochen Wang, Huichi Zhou, Qihui Zhang, Pan Zhou, Yao Wan, and Lichao Sun.
\newblock Mllm-as-a-judge: Assessing multimodal llm-as-a-judge with vision-language benchmark.
\newblock \emph{arXiv preprint arXiv:2402.04788}, 2024{\natexlab{a}}.

\bibitem[Chen et~al.(2024{\natexlab{b}})Chen, Huang, Wu, Tang, Chen, Bai, He, Wang, Zhou, Li, et~al.]{chen2024gui}
Dongping Chen, Yue Huang, Siyuan Wu, Jingyu Tang, Liuyi Chen, Yilin Bai, Zhigang He, Chenlong Wang, Huichi Zhou, Yiqiang Li, et~al.
\newblock Gui-world: A dataset for gui-oriented multimodal llm-based agents.
\newblock \emph{arXiv preprint arXiv:2406.10819}, 2024{\natexlab{b}}.

\bibitem[Chen et~al.(2024{\natexlab{c}})Chen, Li, Dong, Zhang, Zang, Chen, Duan, Wang, et~al.]{chen2024we}
Lin Chen, Jinsong Li, Xiaoyi Dong, Pan Zhang, Yuhang Zang, Zehui Chen, Haodong Duan, Jiaqi Wang, et~al.
\newblock Are we on the right way for evaluating large vision-language models?
\newblock \emph{arXiv preprint arXiv:2403.20330}, 2024{\natexlab{c}}.

\bibitem[Chen et~al.(2024{\natexlab{d}})Chen, Du, Wen, Zhou, Cui, Weng, Tu, Wang, Tong, HUANG, et~al.]{chen2024mj}
Zhaorun Chen, Yichao Du, Zichen Wen, Yiyang Zhou, Chenhang Cui, Zhenzhen Weng, Haoqin Tu, Chaoqi Wang, Zhengwei Tong, Leria HUANG, et~al.
\newblock Mj-bench: Is your multimodal reward model really a good judge?
\newblock In \emph{ICML 2024 Workshop on Foundation Models in the Wild}, 2024{\natexlab{d}}.

\bibitem[Chiang et~al.(2024)Chiang, Zheng, Sheng, Angelopoulos, Li, Li, Zhang, Zhu, Jordan, Gonzalez, and Stoica]{chiang2024chatbot}
Wei-Lin Chiang, Lianmin Zheng, Ying Sheng, Anastasios~Nikolas Angelopoulos, Tianle Li, Dacheng Li, Hao Zhang, Banghua Zhu, Michael Jordan, Joseph~E. Gonzalez, and Ion Stoica.
\newblock Chatbot arena: An open platform for evaluating llms by human preference, 2024.

\bibitem[Chu et~al.(2024)Chu, Ai, Tu, Li, and Liu]{chu2024pre}
Zhumin Chu, Qingyao Ai, Yiteng Tu, Haitao Li, and Yiqun Liu.
\newblock Pre: A peer review based large language model evaluator.
\newblock \emph{arXiv preprint arXiv:2401.15641}, 2024.

\bibitem[Cohen et~al.(2009)Cohen, Huang, Chen, Benesty, Benesty, Chen, Huang, and Cohen]{cohen2009pearson}
Israel Cohen, Yiteng Huang, Jingdong Chen, Jacob Benesty, Jacob Benesty, Jingdong Chen, Yiteng Huang, and Israel Cohen.
\newblock Pearson correlation coefficient.
\newblock \emph{Noise reduction in speech processing}, pages 1--4, 2009.

\bibitem[Deng et~al.(2023)Deng, Zhao, Tang, Gerstein, and Cohan]{deng2023benchmark}
Chunyuan Deng, Yilun Zhao, Xiangru Tang, Mark Gerstein, and Arman Cohan.
\newblock Benchmark probing: Investigating data leakage in large language models.
\newblock In \emph{NeurIPS 2023 Workshop on Backdoors in Deep Learning-The Good, the Bad, and the Ugly}, 2023.

\bibitem[Driess et~al.(2023)Driess, Xia, Sajjadi, Lynch, Chowdhery, Ichter, Wahid, Tompson, Vuong, Yu, et~al.]{driess2023palm}
Danny Driess, Fei Xia, Mehdi~SM Sajjadi, Corey Lynch, Aakanksha Chowdhery, Brian Ichter, Ayzaan Wahid, Jonathan Tompson, Quan Vuong, Tianhe Yu, et~al.
\newblock Palm-e: An embodied multimodal language model.
\newblock \emph{arXiv preprint arXiv:2303.03378}, 2023.

\bibitem[Gao et~al.(2024)Gao, Zhang, Chen, Huang, Wu, Fu, Wan, Zhang, and Sun]{gao2024best}
Chujie Gao, Qihui Zhang, Dongping Chen, Yue Huang, Siyuan Wu, Zhengyan Fu, Yao Wan, Xiangliang Zhang, and Lichao Sun.
\newblock The best of both worlds: Toward an honest and helpful large language model.
\newblock \emph{arXiv preprint arXiv:2406.00380}, 2024.

\bibitem[Hong et~al.(2024)Hong, Wang, Lv, Xu, Yu, Ji, Wang, Dong, et~al.]{hong2024cogagent}
Wenyi Hong, Weihan Wang, Qingsong Lv, Jiazheng Xu, Wenmeng Yu, Junhui Ji, Yan Wang, Yuxiao Dong, et~al.
\newblock Cogagent: A visual language model for gui agents.
\newblock In \emph{Proceedings of the IEEE/CVF Conference on Computer Vision and Pattern Recognition}, pages 14281--14290, 2024.

\bibitem[Kasner and Du{\v{s}}ek(2024)]{kasner2024beyond}
Zden{\v{e}}k Kasner and Ond{\v{r}}ej Du{\v{s}}ek.
\newblock Beyond traditional benchmarks: Analyzing behaviors of open llms on data-to-text generation.
\newblock In \emph{Proceedings of the 62nd Annual Meeting of the Association for Computational Linguistics (Volume 1: Long Papers)}, pages 12045--12072, 2024.

\bibitem[Kasner and Dušek(2024)]{kasner2024traditionalbenchmarksanalyzingbehaviors}
Zdeněk Kasner and Ondřej Dušek.
\newblock Beyond traditional benchmarks: Analyzing behaviors of open llms on data-to-text generation, 2024.

\bibitem[Lab and etc.(2024)]{opensoraplan2024yuan}
PKU-Yuan Lab and Tuzhan~AI etc.
\newblock Open-sora-plan, 2024.

\bibitem[LastName(2014{\natexlab{a}})]{Authors14}
FirstName LastName.
\newblock The frobnicatable foo filter, 2014{\natexlab{a}}.
\newblock Face and Gesture submission ID 324. Supplied as supplemental material {\tt fg324.pdf}.

\bibitem[LastName(2014{\natexlab{b}})]{Authors14b}
FirstName LastName.
\newblock Frobnication tutorial, 2014{\natexlab{b}}.
\newblock Supplied as supplemental material {\tt tr.pdf}.

\bibitem[Li and Lu(2024)]{li2024survey}
Jian Li and Weiheng Lu.
\newblock A survey on benchmarks of multimodal large language models.
\newblock \emph{arXiv preprint arXiv:2408.08632}, 2024.

\bibitem[Li et~al.(2023)Li, Patel, and Du]{li2023prd}
Ruosen Li, Teerth Patel, and Xinya Du.
\newblock Prd: Peer rank and discussion improve large language model based evaluations.
\newblock \emph{arXiv preprint arXiv:2307.02762}, 2023.

\bibitem[Lin et~al.(2023)Lin, Ye, Zhu, Cui, Ning, Jin, and Yuan]{lin2023video}
Bin Lin, Yang Ye, Bin Zhu, Jiaxi Cui, Munan Ning, Peng Jin, and Li Yuan.
\newblock Video-llava: Learning united visual representation by alignment before projection.
\newblock \emph{arXiv preprint arXiv:2311.10122}, 2023.

\bibitem[Liu et~al.(2024{\natexlab{a}})Liu, Li, Wu, and Lee]{liu2024visual}
Haotian Liu, Chunyuan Li, Qingyang Wu, and Yong~Jae Lee.
\newblock Visual instruction tuning.
\newblock \emph{Advances in neural information processing systems}, 36, 2024{\natexlab{a}}.

\bibitem[Liu et~al.(2023)Liu, Lei, Wang, Huang, Feng, Wen, Cheng, Ke, et~al.]{liu2023alignbenchbenchmarkingchinesealignment}
Xiao Liu, Xuanyu Lei, Shengyuan Wang, Yue Huang, Zhuoer Feng, Bosi Wen, Jiale Cheng, Pei Ke, et~al.
\newblock Alignbench: Benchmarking chinese alignment of large language models.
\newblock \emph{arXiv preprint arXiv:2311.18743}, 2023.

\bibitem[Liu et~al.(2024{\natexlab{b}})Liu, Zhou, Guo, Shareghi, Vuli{\'c}, Korhonen, and Collier]{liu2024aligning}
Yinhong Liu, Han Zhou, Zhijiang Guo, Ehsan Shareghi, Ivan Vuli{\'c}, Anna Korhonen, and Nigel Collier.
\newblock Aligning with human judgement: The role of pairwise preference in large language model evaluators.
\newblock \emph{arXiv preprint arXiv:2403.16950}, 2024{\natexlab{b}}.

\bibitem[Liusie et~al.(2023)Liusie, Manakul, and Gales]{liusie2023zero}
Adian Liusie, Potsawee Manakul, and Mark~JF Gales.
\newblock Zero-shot nlg evaluation through pairware comparisons with llms.
\newblock \emph{arXiv preprint arXiv:2307.07889}, 2023.

\bibitem[Lu et~al.(2022)Lu, Mishra, Xia, Qiu, Chang, Zhu, Tafjord, Clark, and Kalyan]{lu2022learn}
Pan Lu, Swaroop Mishra, Tanglin Xia, Liang Qiu, Kai-Wei Chang, Song-Chun Zhu, Oyvind Tafjord, Peter Clark, and Ashwin Kalyan.
\newblock Learn to explain: Multimodal reasoning via thought chains for science question answering.
\newblock \emph{Advances in Neural Information Processing Systems}, 35:\penalty0 2507--2521, 2022.

\bibitem[Meta(2024)]{meta2024llama3.2_11b}
Meta.
\newblock Llama 3.2 11b-vision-instruct.
\newblock \url{https://huggingface.co/meta-llama/Llama-3.2-11B-Vision-Instruct}, 2024.

\bibitem[Moell(2024)]{moell2024evaluating}
Birger Moell.
\newblock Evaluating large language models with human feedback: Establishing a swedish benchmark.
\newblock \emph{arXiv preprint arXiv:2405.14006}, 2024.

\bibitem[Ning et~al.(2024)Ning, Yang, Liu, Yao, Liu, Wang, Pang, and Yuan]{ning2024peer}
Kun-Peng Ning, Shuo Yang, Yu-Yang Liu, Jia-Yu Yao, Zhen-Hui Liu, Yu Wang, Ming Pang, and Li Yuan.
\newblock Peer-review-in-llms: Automatic evaluation method for llms in open-environment.
\newblock \emph{arXiv preprint arXiv:2402.01830}, 2024.

\bibitem[OpenAI(2023)]{openai2023gpt4}
OpenAI.
\newblock Gpt-4 technical report.
\newblock \url{https://openai.com/research/gpt-4}, 2023.

\bibitem[OpenAI(2024{\natexlab{a}})]{openai2024gpt4omini}
OpenAI.
\newblock Gpt-4o mini: Advancing cost-efficient intelligence.
\newblock \url{https://openai.com/index/gpt-4o-mini-advancing-cost-efficient-intelligence/}, 2024{\natexlab{a}}.

\bibitem[OpenAI(2024{\natexlab{b}})]{openai2024o1}
OpenAI.
\newblock Introducing openai o1: Advancing model reasoning with visual and logical insights.
\newblock https://openai.com/index/introducing-openai-o1-preview/, 2024{\natexlab{b}}.

\bibitem[OpenAI(2024{\natexlab{c}})]{openai_gpt4o_2024}
OpenAI.
\newblock Hello gpt-4o, 2024{\natexlab{c}}.

\bibitem[Pi et~al.(2024)Pi, Yao, Gao, Zhang, and Zhang]{pi2024perceptiongpt}
Renjie Pi, Lewei Yao, Jiahui Gao, Jipeng Zhang, and Tong Zhang.
\newblock Perceptiongpt: Effectively fusing visual perception into llm.
\newblock In \emph{Proceedings of the IEEE/CVF Conference on Computer Vision and Pattern Recognition}, pages 27124--27133, 2024.

\bibitem[Radford et~al.(2021)Radford, Kim, Hallacy, Ramesh, Goh, Agarwal, Sastry, Askell, Mishkin, Clark, et~al.]{radford2021learning}
Alec Radford, Jong~Wook Kim, Chris Hallacy, Aditya Ramesh, Gabriel Goh, Sandhini Agarwal, Girish Sastry, Amanda Askell, Pamela Mishkin, Jack Clark, et~al.
\newblock Learning transferable visual models from natural language supervision.
\newblock In \emph{International conference on machine learning}, pages 8748--8763. PMLR, 2021.

\bibitem[Raina et~al.(2024)Raina, Liusie, and Gales]{raina2024llm}
Vyas Raina, Adian Liusie, and Mark Gales.
\newblock Is llm-as-a-judge robust? investigating universal adversarial attacks on zero-shot llm assessment.
\newblock \emph{arXiv preprint arXiv:2402.14016}, 2024.

\bibitem[Raju et~al.(2024)Raju, Jain, Li, Li, and Thakkar]{raju2024constructing}
Ravi Raju, Swayambhoo Jain, Bo Li, Jonathan Li, and Urmish Thakkar.
\newblock Constructing domain-specific evaluation sets for llm-as-a-judge.
\newblock \emph{arXiv preprint arXiv:2408.08808}, 2024.

\bibitem[Refuel.ai(2025)]{refuel2025llmlabeling}
Refuel.ai.
\newblock Llm labeling: Technical report, 2025.

\bibitem[Roziere et~al.(2023)Roziere, Gehring, Gloeckle, Sootla, Gat, Tan, Adi, Liu, Sauvestre, Remez, et~al.]{roziere2023code}
Baptiste Roziere, Jonas Gehring, Fabian Gloeckle, Sten Sootla, Itai Gat, Xiaoqing~Ellen Tan, Yossi Adi, Jingyu Liu, Romain Sauvestre, Tal Remez, et~al.
\newblock Code llama: Open foundation models for code.
\newblock \emph{arXiv preprint arXiv:2308.12950}, 2023.

\bibitem[Spearman(1961)]{spearman1961proof}
Charles Spearman.
\newblock The proof and measurement of association between two things.
\newblock 1961.

\bibitem[Stephan et~al.(2024)Stephan, Zhu, A{\ss}enmacher, Shen, and Roth]{stephan2024calculation}
Andreas Stephan, Dawei Zhu, Matthias A{\ss}enmacher, Xiaoyu Shen, and Benjamin Roth.
\newblock From calculation to adjudication: Examining llm judges on mathematical reasoning tasks.
\newblock \emph{arXiv preprint arXiv:2409.04168}, 2024.

\bibitem[Surowiecki(2005)]{surowiecki2005wisdom}
James Surowiecki.
\newblock The wisdom of crowds/james surowiecki.
\newblock \emph{NY.: Anchor}, 2005.

\bibitem[Tan et~al.(2024)Tan, Zhuang, Montgomery, Tang, Cuadron, Wang, Popa, and Stoica]{tan2024judgebench}
Sijun Tan, Siyuan Zhuang, Kyle Montgomery, William~Y Tang, Alejandro Cuadron, Chenguang Wang, Raluca~Ada Popa, and Ion Stoica.
\newblock Judgebench: A benchmark for evaluating llm-based judges.
\newblock \emph{arXiv preprint arXiv:2410.12784}, 2024.

\bibitem[Tang et~al.(2020)Tang, Ma, Zhang, Wu, and Yang]{tang2020semantic}
Ruixue Tang, Chao Ma, Wei~Emma Zhang, Qi Wu, and Xiaokang Yang.
\newblock Semantic equivalent adversarial data augmentation for visual question answering.
\newblock In \emph{Computer Vision--ECCV 2020: 16th European Conference, Glasgow, UK, August 23--28, 2020, Proceedings, Part XIX 16}, pages 437--453. Springer, 2020.

\bibitem[Team et~al.(2023)Team, Anil, Borgeaud, Wu, Alayrac, Yu, Soricut, et~al.]{team2023gemini}
Gemini Team, Rohan Anil, Sebastian Borgeaud, Yonghui Wu, Jean-Baptiste Alayrac, Jiahui Yu, Radu Soricut, et~al.
\newblock Gemini: a family of highly capable multimodal models.
\newblock \emph{arXiv preprint arXiv:2312.11805}, 2023.

\bibitem[Wang et~al.(2024)Wang, Zhang, and Patel]{wang2024userbenchmark}
Li Wang, Hui Zhang, and Rajesh Patel.
\newblock A user-centric benchmark for evaluating large language models.
\newblock \emph{arXiv preprint arXiv:2404.13940}, 2024.

\bibitem[Wei et~al.(2022)Wei, Wang, Schuurmans, Bosma, Xia, Chi, Le, Zhou, et~al.]{wei2022chain}
Jason Wei, Xuezhi Wang, Dale Schuurmans, Maarten Bosma, Fei Xia, Ed Chi, Quoc~V Le, Denny Zhou, et~al.
\newblock Chain-of-thought prompting elicits reasoning in large language models.
\newblock \emph{Advances in neural information processing systems}, 35:\penalty0 24824--24837, 2022.

\bibitem[Wu et~al.(2023)Wu, Fei, Qu, Ji, and Chua]{wu2023next}
Shengqiong Wu, Hao Fei, Leigang Qu, Wei Ji, and Tat-Seng Chua.
\newblock Next-gpt: Any-to-any multimodal llm.
\newblock \emph{arXiv preprint arXiv:2309.05519}, 2023.

\bibitem[Wu et~al.(2024)Wu, Huang, Gao, Chen, Zhang, Wan, Zhou, Zhang, Gao, Xiao, et~al.]{wu2024unigen}
Siyuan Wu, Yue Huang, Chujie Gao, Dongping Chen, Qihui Zhang, Yao Wan, Tianyi Zhou, Xiangliang Zhang, Jianfeng Gao, Chaowei Xiao, et~al.
\newblock Unigen: A unified framework for textual dataset generation using large language models.
\newblock \emph{arXiv preprint arXiv:2406.18966}, 2024.

\bibitem[xAI(2024)]{RealWorldQA}
xAI.
\newblock Realworldqa: A benchmark for evaluating real-world spatial understanding in multimodal ai models, 2024.

\bibitem[Xu et~al.(2024)Xu, Wang, Fan, and Liu]{xu2024benchmarking}
Ruijie Xu, Zengzhi Wang, Run-Ze Fan, and Pengfei Liu.
\newblock Benchmarking benchmark leakage in large language models.
\newblock \emph{arXiv preprint arXiv:2404.18824}, 2024.

\bibitem[Yang et~al.(2024)Yang, Teng, Zheng, Ding, Huang, Xu, Yang, Hong, Zhang, Feng, et~al.]{yang2024cogvideox}
Zhuoyi Yang, Jiayan Teng, Wendi Zheng, Ming Ding, Shiyu Huang, Jiazheng Xu, Yuanming Yang, Wenyi Hong, Xiaohan Zhang, Guanyu Feng, et~al.
\newblock Cogvideox: Text-to-video diffusion models with an expert transformer.
\newblock \emph{arXiv preprint arXiv:2408.06072}, 2024.

\bibitem[Ye et~al.(2024)Ye, Wang, Huang, Chen, Zhang, Moniz, Gao, Geyer, Huang, Chen, et~al.]{ye2024justice}
Jiayi Ye, Yanbo Wang, Yue Huang, Dongping Chen, Qihui Zhang, Nuno Moniz, Tian Gao, Werner Geyer, Chao Huang, Pin-Yu Chen, et~al.
\newblock Justice or prejudice? quantifying biases in llm-as-a-judge.
\newblock \emph{arXiv preprint arXiv:2410.02736}, 2024.

\bibitem[Yu et~al.(2023)Yu, Yang, Li, Wang, Lin, Liu, Wang, and Wang]{yu2023mm}
Weihao Yu, Zhengyuan Yang, Linjie Li, Jianfeng Wang, Kevin Lin, Zicheng Liu, Xinchao Wang, and Lijuan Wang.
\newblock Mm-vet: Evaluating large multimodal models for integrated capabilities.
\newblock \emph{arXiv preprint arXiv:2308.02490}, 2023.

\bibitem[Yuan et~al.(2024{\natexlab{a}})Yuan, Huang, Shi, Xu, Zhu, Lin, Cheng, Yuan, and Luo]{yuan2024magictime}
Shenghai Yuan, Jinfa Huang, Yujun Shi, Yongqi Xu, Ruijie Zhu, Bin Lin, Xinhua Cheng, Li Yuan, and Jiebo Luo.
\newblock Magictime: Time-lapse video generation models as metamorphic simulators.
\newblock \emph{arXiv preprint arXiv:2404.05014}, 2024{\natexlab{a}}.

\bibitem[Yuan et~al.(2024{\natexlab{b}})Yuan, Huang, Xu, Liu, Zhang, Shi, Zhu, Cheng, Luo, and Yuan]{yuan2024chronomagic}
Shenghai Yuan, Jinfa Huang, Yongqi Xu, Yaoyang Liu, Shaofeng Zhang, Yujun Shi, Ruijie Zhu, Xinhua Cheng, Jiebo Luo, and Li Yuan.
\newblock Chronomagic-bench: A benchmark for metamorphic evaluation of text-to-time-lapse video generation.
\newblock \emph{arXiv preprint arXiv:2406.18522}, 2024{\natexlab{b}}.

\bibitem[Yue et~al.(2024)Yue, Ni, Zhang, Zheng, Liu, Zhang, Stevens, Jiang, et~al.]{yue2024mmmu}
Xiang Yue, Yuansheng Ni, Kai Zhang, Tianyu Zheng, Ruoqi Liu, Ge Zhang, Samuel Stevens, Dongfu Jiang, et~al.
\newblock Mmmu: A massive multi-discipline multimodal understanding and reasoning benchmark for expert agi.
\newblock In \emph{Proceedings of the IEEE/CVF Conference on Computer Vision and Pattern Recognition}, pages 9556--9567, 2024.

\bibitem[Zar(2005)]{zar2005spearman}
Jerrold~H Zar.
\newblock Spearman rank correlation.
\newblock \emph{Encyclopedia of biostatistics}, 7, 2005.

\bibitem[Zheng et~al.(2023)Zheng, Chiang, Sheng, Zhuang, Wu, Zhuang, Lin, Li, Li, Xing, et~al.]{zheng2023judging}
Lianmin Zheng, Wei-Lin Chiang, Ying Sheng, Siyuan Zhuang, Zhanghao Wu, Yonghao Zhuang, Zi Lin, Zhuohan Li, Dacheng Li, Eric Xing, et~al.
\newblock Judging llm-as-a-judge with mt-bench and chatbot arena.
\newblock \emph{Advances in Neural Information Processing Systems}, 36:\penalty0 46595--46623, 2023.

\bibitem[Zhu et~al.(2023)Zhu, Lin, Ning, Yan, Cui, Wang, Pang, Jiang, Zhang, Li, et~al.]{zhu2023languagebind}
Bin Zhu, Bin Lin, Munan Ning, Yang Yan, Jiaxi Cui, HongFa Wang, Yatian Pang, Wenhao Jiang, Junwu Zhang, Zongwei Li, et~al.
\newblock Languagebind: Extending video-language pretraining to n-modality by language-based semantic alignment.
\newblock \emph{arXiv preprint arXiv:2310.01852}, 2023.

\end{thebibliography}
}

\clearpage
\setcounter{page}{1}
\maketitlesupplementary

\section{Model Selection}
We select models as follows:

\textbf{GPT-4o} \cite{openai_gpt4o_2024} A versatile multimodal model by OpenAI, handling text, image, and audio inputs. It excels in vision and language tasks with enhanced processing speed. Known for strong real-time performance in audio and vision, GPT-4o is ideal for a variety of applications, including multilingual tasks.

\textbf{GPT-4o mini} \cite{openai2024gpt4omini} A smaller, cost-effective version of GPT-4o, optimized for handling text and images, with plans for audio support. It is designed for high-volume, real-time applications like chatbots and coding tasks, offering strong performance at a lower cost.

\textbf{Gemini-1.5-Pro} \cite{team2023gemini} Developed by Google DeepMind, this model uses Mixture-of-Experts architecture to optimize performance. It supports up to 1 million tokens and excels in translation, coding, and multimodal tasks. It is ideal for enterprise use due to its cost-efficiency and scalability.

\textbf{Claude-3.5-Sonnet} \cite{anthropic2024claude35} From Anthropic, this model is optimized for reasoning, coding, and multimodal tasks. It excels in complex problem-solving and visual understanding, making it useful for customer support and detailed code-generation tasks.

\textbf{Claude-3-Haiku} \cite{anthropic2024claude3haiku} Developed by Anthropic, Claude 3.5 Haiku is a high-speed language model optimized for rapid response and advanced reasoning. With a 200K token context window and a maximum output of 4,096 tokens, it efficiently handles large datasets. Its affordability and speed make it ideal for applications requiring quick, concise responses, such as interactive chatbots and real-time data analysis.

\textbf{Llama-3.2-11B-Vision-instruct} \cite{meta2024llama3.2_11b} a multimodal large language model from Meta with 11 billion parameters, designed to handle both text and image inputs. It excels in tasks such as image captioning, visual question answering, and interpreting complex visual data. This model is particularly effective for industries like healthcare and retail, where real-time visual and textual analysis is key.

\section{Extended Experiment}

\subsection{Pre-experiment for Different Weights.}
\label{differentW}

We have conducted a preliminary experiment to investigate the relationship between the confidence weights and scores of models, as a substantiation of our methodology. Below is a detailed description of the experiment and its findings, which align with the methodology discussed in the paper.

To begin, we designed a toy experiment to examine the role of confidence weights \( w \). Based on the scores on manually designed benchmarks, we can obtain a model score ranking list, then designing weight configurations that are either consistent or reverse to this score ranking. Specifically, we constructed three weighting methods: Reverse Weight (\( w = [0, 0.1, \dots, 1] \)), Uniform Weight (\( w = [1, 1, \dots, 1] \)), and Consistent Weight (\( w = [1, 0.9, \dots, 0] \)). Using these manually constructed weight configurations, we calculated the response score \( G_j \) for each model based on the predefined \autoref{eq6} in \autoref{method} and obtained the score list \( \hat{G} \) for all models. We then measured the alignment between the obtained ranking \( \hat{G} \) and the human-annotated score list \( G \) using predefined metrics.

The results as summarized in \autoref{tab:pre_ex}, demonstrate that the Consistent Weight configuration achieves the highest correlation values, while the Reverse Weight configuration consistently yields the poorest results. These findings validate the proposed consistency assumption: assigning higher weights to models with stronger capabilities leads to better alignment between the model score list and the human-annotated one. In essence, responses recognized more favorably by other “reviewers” (models) tend to originate from higher-level models. This reinforces the idea that high-capability MLLMs evaluate others’ responses more accurately and achieve higher scores.

We formalize this observation as the consistency assumption, which states that: 1. High-level LLMs exhibit greater confidence and accuracy when evaluating responses compared to lower-level ones. 2. A model’s ability and its associated score are generally consistent.

Building on this preliminary finding, we devised an optimization framework aimed at maximizing the consistency between each model’s capability (\( w \)) and its response score (\( G \)), constrained by our proposed methodology.

\begin{table}[t]
    \centering
    \setlength\tabcolsep{1.0mm}
    \renewcommand\arraystretch{1.1}
    \def \mysp {\hspace{7pt}}
    {\small 
    \resizebox{\linewidth}{!}{
    \begin{tabular}{ccccc}
    \toprule[1.5pt]
    \multirow{2}{*}{{\makecell{Method}}} & \multicolumn{2}{c}{MMstar} & \multicolumn{2}{c}{ScienceQA} \\
    \cmidrule(lr){2-3} \cmidrule(lr){4-5}
          & \textbf{Pearson} & \textbf{Spearman} & \textbf{Pearson} & \textbf{Spearman} \\
    \hline
        \makecell{Reverse Weight} & 0.607 & 0.486 & 0.334 & 0.257 \\

        \makecell{Uniform Weight} & 0.725 & 0.771 & 0.463 & 0.686 \\ 

        {\makecell{Consistent Weight}} & 0.807 & 0.829 & 0.760 & 0.771 \\

    \hline
    $\textbf{UPME}$  & \textbf{0.944} & \textbf{0.972} & \textbf{0.814} & \textbf{0.886} \\
    \bottomrule[1.5pt]
    \end{tabular}}
    }
    \caption{Performance comparison of Consistent, Uniform and Reverse weight.}
    \label{tab:pre_ex}

\end{table}

\subsection{More Datasets}
We have experimented on MMVet and the results in \autoref{tab:PRD} show that UPME maintains its superior performance.

\begin{table}[ht]
    \vspace{-10pt}
    {\small
    \centering
    \resizebox{\linewidth}{!}{
    \begin{tabular}{l|cc|cc|cc}
    \toprule[1.5pt]
    \multicolumn{1}{c|}{\multirow{2}*{\textbf{Models}}} & \multicolumn{2}{c|}{\textbf{MMstar}} & \multicolumn{2}{c|}{\textbf{ScienceQA}} & \multicolumn{2}{c}{\textbf{MMVet}}\\
     & Pearson & Spearman & Pearson & Spearman & Pearson & Spearman\\
    \midrule
    Peer Review & 0.725 & 0.771 & 0.463 & 0.686 & 0.688 & 0.752 \\
    Majority Vote  & 0.757 & 0.757 & 0.509 & 0.524 & 0.732 & 0.643 \\
    Rating Vote & 0.795 & 0.743 & 0.623 & 0.629 & 0.739 & 0.743\\
    PRD \cite{li2023prd} & 0.838 & 0.864 & 0.692 & 0.636 & 0.794 & 0.814\\
    UPME & 0.944 & 0.972 & 0.814 & 0.886 & 0.914 & 0.928\\
    \bottomrule[1.5pt]
    \end{tabular}}

    \caption{\small Comparison with recent methods.}
    \label{tab:PRD}
    }

\end{table}

\subsection{Hyperparameter}

The weights $\gamma_1$, $\gamma_2$, and $\gamma_3$ in \autoref{eq9} were initialized as 0.4, 0.4, and 0.2, respectively, reflecting a balanced emphasis on response correctness and visual understanding while slightly de-emphasizing image-text correlation. This choice is intuited that correctness and reasoning typically have a larger impact on multimodal evaluation tasks.

\begin{table}[ht]
  \centering
  \resizebox{0.6\linewidth}{!}{
    \begin{tabular}{cccccc}
    \toprule[1.5pt]
         $\bm{\gamma_1}$ & $\bm{\gamma_2}$ & $\bm{\gamma_3}$ & \textbf{Pearson} & \textbf{Spearman} \\
         \midrule
            0.4 & 0.4 & 0.2 & 0.9415 & 0.9441 \\
            0.3 & 0.3 & 0.4 & 0.9397 & 0.8581 \\
            0.5 & 0.3 & 0.2 & 0.9306 & 0.7174 \\
            0.3 & 0.5 & 0.2 & 0.9365 & 0.8857 \\
    \bottomrule[1.5pt]
    \end{tabular}
  }
  % \vspace{-5pt}
  \caption{hyperparameter in Scoring criteria.}
  \label{tab:hyperparameter}
\end{table}

To validate the optimality of this combination, we conducted experiments with different hyperparameter configurations. The results for four representative settings are summarized in Table \autoref{tab:hyperparameter}. The proposed setting achieves the highest Pearson and Spearman correlations, indicating its effectiveness in aligning with human evaluations.

Notably, our experiments also show that the framework is relatively insensitive to small variations in these weights within the range [0.2,0.5], demonstrating its robustness.
\vspace{3pt}

\textbf{Task-Specific Flexibility: }The UPME framework supports task-specific flexibility. For instance, users may adjust $\gamma_3$ to prioritize image-text correlation in tasks requiring strong alignment between modalities or increase $\gamma_2$ for tasks demanding advanced reasoning capabilities, which allows the framework to cater to diverse evaluation needs.

\textbf{Future Directions: }While manual tuning of hyperparameters has proven effective, we agree that automating this process would further enhance the framework’s generality and ease of use. We are actively exploring automated methods, such as validation-based optimization techniques or reinforcement learning approaches, to dynamically determine these weights based on task characteristics.

\subsection{Reliability of Judge Correctness}

\textbf{Advantages of UPME’s Question Generation: }In the original peer review mechanism, the review model might encounter questions that it cannot answer accurately, leading to unreliable evaluations. In contrast, the UPME framework enables the review model to generate questions autonomously, ensuring that these questions fall within its capability. This significantly enhances the reliability of the review model in assessing the correctness of responses from the evaluated models.

\textbf{Empirical Evidence of Reliability: }As shown in  \autoref{tab:JudgeCorrect}, UPME demonstrates a substantial improvement in both accuracy and human agreement compared to the original peer review mechanism.

\begin{table}[ht]
      \centering
      \resizebox{0.8\linewidth}{!}{
        \begin{tabular}{lccc}
        \toprule[1.5pt]
        \textbf{Method} & \textbf{Dataset} & \textbf{Accuracy (\%)}  & \makecell{\textbf{Human (\%)} \\ \textbf{Alignment} }\\
        \midrule
    \multirow{2}{*}{{\makecell{Peer \\ Review}}} & MMStar & 64.2 & 71.1 \\
      & ScienceQA & 60.3 & 68.2 \\
    \multirow{2}{*}{UPME} & MMStar & 87.8 & 95.9 \\
     & ScienceQA & 79.6 & 87.4 \\
        \bottomrule[1.5pt]
        \end{tabular}
      }
      % \vspace{-5pt}
      \caption{ $Judge_{Correct}$ reliability.}
      \label{tab:JudgeCorrect}
\end{table}

\section{More Information about UPME}

\subsection{The Cost of UPME}

UPME significantly reduces both time and financial costs: \textbf{Time Costs:} Creating VQAs manually requires deep understanding and may take several minutes to hours per task. AI tools significantly reduce this time, with labeling efficiency improved by up to 100 times~\cite{refuel2025llmlabeling}. UPME further accelerates evaluation, processing dozens of images per second. \textbf{Financial Costs:} Human annotations cost 1–5 per image depending on complexity, while UPME reduces this to approximately 1/7 of the manual cost.

Baseline methods like PeerReview and Majority Vote require extensive human-labeled data, significantly increasing time and costs. UPME eliminates manual annotation, offering efficient evaluations.

\begin{table}[ht]
\vspace{-5pt}

\centering
\small
\renewcommand\arraystretch{1.1}
\resizebox{0.9\linewidth}{!}{
\begin{tabular}{lcc} \toprule[1.5pt] 
\textbf{Method} & \textbf{Time / img} & \textbf{Finance / img} \\ 
\hline 
Human Annotation & 3$\sim$10 min  & \$ 2$\sim$7 \\
Majority Vote & \multirow{2}{*}{{2$\sim$8 min}} & \multirow{2}{*}{{\$ 1$\sim$5}} \\
Rating Vote & & \\
UPME Framework & 1.5 s & \$ 0.15 \\
\bottomrule[1.5pt] \end{tabular} 
}
% \vspace{-5pt}
\caption{Comparison of Time and Financial Costs}
\label{tab:cost}
% \vspace{-10pt}
\end{table}

\subsection{Image-Text Correlation}
\label{detailsofClip}

To compute the Image-Text Correlation score \( S_{\text{Clip}} \), we employ the CLIP model \cite{radford2021learning}, which measures the cosine similarity between image embeddings and text embeddings. For textual responses \( A_i^{j,r} \) exceeding CLIP’s maximum token input limit of 77 tokens, we implement a segmentation strategy that ensures each segment contains no more than the limit, preserving the context across the text. 

Specifically, if the number of tokens \( n \) in a response exceeds 77, we calculate the starting indices for each segment by dividing the range from \( 0 \) to \( n - 77 \) into five equal intervals. These numbers serve as the starting points for each segment. Each segment then extends for 77 tokens from its starting index, ensuring full coverage of the original text with some overlap between consecutive segments. This overlapping is crucial as it helps preserve the continuity and context of the text, which might otherwise be lost if the segments were disjointed. The segments are then processed alongside the image through the CLIP model to generate embeddings, and cosine similarities between each text segment and the image are calculated. We derive the average of these similarity scores to evaluate the text-image alignment.

A notable feature of the segmentation strategy addresses potential verbosity bias by penalizing segments containing irrelevant or poorly aligned content. By computing the average cosine similarity across all segments, the approach inherently discounts segments that introduce irrelevant or poorly aligned content, reducing the score for long but less relevant responses. This mechanism effectively counteracts the verbosity bias of MLLM-as-a-judge.

\subsection{Algorithm of UPME}

\begin{algorithm}[h]
    \caption{Algorithm of UPME}
    \label{alg: UPME}
    \renewcommand{\algorithmicrequire}{\textbf{Input:}}
    \renewcommand{\algorithmicensure}{\textbf{Output:}}
    \begin{algorithmic}[1]
    
        \Require MLLM Pool $\mathcal{M}$, Image pool $\mathcal{I}$, Epochs $T$
        
        \Ensure Model Scores $\hat{G}$, Weights $w$
        
        \State Initialize $w$ and $G$ for models in $M$
        \State \textbf{// Dynamic Update $\hat{G}$ and $w$}
        \For {each iteration $t = 1$ to $T$}

            \State Randomly select $M_r, M_j, M_k \in M$
            \State Generate $Q_i^r, A_i^{j,r}, A_i^{k,r}$
            
            \State Calculate $S_{VL}(A_i^{j,r}, A_i^{k,r}, Q_i^r, I_i | M_r)$
            
            \State Update scores using EMA:
            \State $G[M_j] \leftarrow (1 - \alpha) G[M_j] + \alpha S_{VL}$

            \State $w \leftarrow \text{optimize\_weights}(G)$
        \EndFor
        
    \end{algorithmic}
\end{algorithm}

\subsection{Human Preference Alignment}
\label{humanannotation}

\begin{figure}
    \centering
    \includegraphics[width=0.95\linewidth]{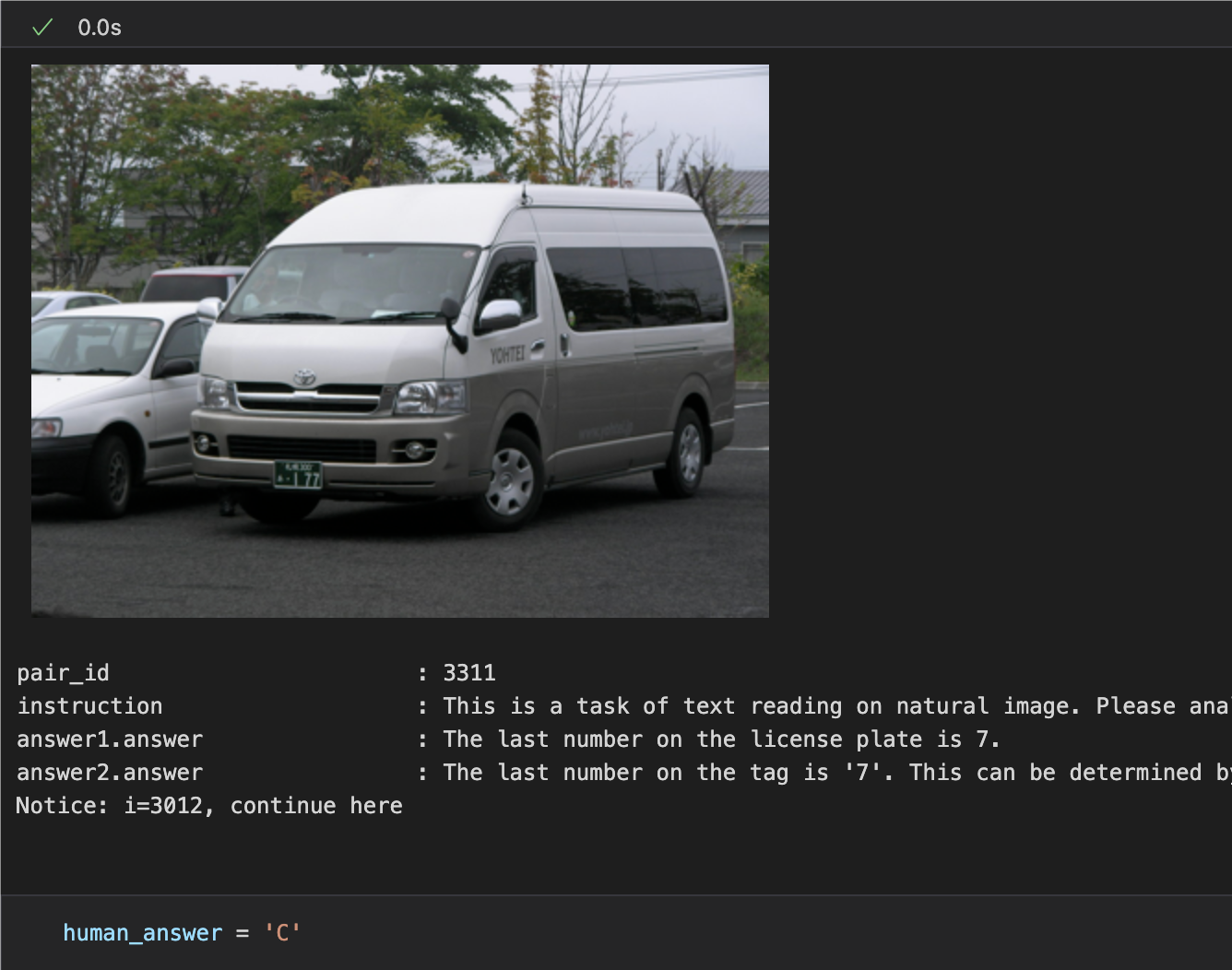}
    \caption{Screenshot of human annotation.}
    \vspace{5pt}
    \label{screenshot}
\end{figure}

The annotation work for human preferences alignment was carried out by five human experts with professional English proficiency, taking them a total of 170 hours. The labeling screenshot is shown in \autoref{screenshot}. The guidelines for human annotation are shown in \autoref{guideline}. Each data is associated with an image, a review model, and two candidate models, and it requires the completion of the following two annotation tasks: 1) Without knowledge of the review model's judgment, the annotator provides their own choice. 2) After being informed of the review model's judgment, the annotator indicates whether they agree with the decision.

These two tasks are assigned to different individuals for the same image, meaning that the same annotator does not perform both tasks for the same image. Each task is annotated by two annotators. When the results of the two annotators are consistent, the image's human preference annotation is obtained. If the results are inconsistent, up to five annotators will vote, and a majority vote determines the final annotation result for the data. Statistical analysis shows that such cases requiring voting account for only 2.17\% of the final annotation results.

The experimental results on \autoref{tab:humanalignment} indicate that the baseline method exhibited relatively low human agreement and model consistency rates, suggesting that the Peer Review mechanism under an unsupervised setting without weight optimization struggles to align with human preferences. In contrast, UPME demonstrated significant improvements by incorporating metrics such as Correctness, Visual Understanding, and Clip Relevance. On the MMstar dataset, UPME achieved an agreement rate of 95.9\% and a consistency rate of 89.8\%, showing that the optimized scoring criteria significantly enhance the accuracy of evaluation outputs and alignment with human preferences. By capturing key multimodal understanding metrics without relying on manual labeling, UPME effectively achieved high consistency with human annotations, highlighting its substantial advantages in improving response consistency and accuracy under an unsupervised framework.

\newpage

\begin{figure*}

    \begin{tcolorbox}[guideline, title=Human annotation guideline]

[Task Overview]\\

You are a human expert tasked with annotating the data assigned to you. You need to evaluate the responses of two candidate models to a given image and question, make your judgment, and assess whether the review model's judgment is correct. Each annotation involves assessing data instances that include an image, the responses from two candidate models, and a judgment from the review model.\\

For the given data, you will perform one of two tasks and focus your assessment on one of two aspects. Please be aware of which task the data belongs to and which aspect of the candidate models' responses you are evaluating.\\

[Two Annotation Tasks]\\

\#\#\# Task 1: Independent Choice Without Review Model Judgment\\
- You should independently evaluate and select your preferred response between the two candidate models based on their response to the image-related question.\\
- No information about the review model's judgment is provided during this step.\\

\#\#\# Task 2: Agreement with Review Model Judgment\\
- You are informed of the review model's judgment and asked to decide whether you agree or disagree with it.\\

Note: Tasks 1 and 2 must be conducted by different individuals for the same image to prevent cognitive bias.\\

[Two Aspects to Evaluate]\\

When evaluating the responses from the two candidate models, you need to focus on one of the following two aspects:\\

\#\#\# 1) Correctness\\
- Your evaluation should be strictly objective, focusing only on which response is correct. Please proceed as follows:\\
  - If only one model provides a correct answer, identify the correct model.\\
  - If both answers are correct or both are incorrect, output 'C' to indicate a tie.\\

\#\#\# 2) Visual Understanding and Reasoning\\
- Focus exclusively on the depth of visual understanding and the quality of reasoning in each response. Do not evaluate based on correctness. Here are the Evaluation Criteria:\\
  - Captioning: Evaluate the ability to generate precise descriptions of image elements.\\
  - Reasoning: Assess logical consistency and coherence in explanations and conclusions.\\
  - Grounding: Evaluate accurate object localization within the image.\\
  - Relationship: Assess the understanding of relationships and interactions between subjects in the image.\\

\end{tcolorbox}
\caption{Human annotation guideline.}
\label{guideline}
\end{figure*}

\clearpage
% \newpage

\section{Prompt Template}
\label{sec:prompt_template}

\begin{figure}[ht]
\centering
\resizebox{\textwidth}{!}{
    \begin{tcolorbox}[prompt, title=Question generation prompt for review model]

You are a review model tasked with evaluating the visual capabilities of two other models.\\

Based on the provided image input, generate a question that is directly related to the content of the image.\\

Please respond with only the question and no additional content.

\end{tcolorbox}
}
\end{figure}

\begin{figure}[ht]
    \centering
    \resizebox{\textwidth}{!}{
    \begin{tcolorbox}[prompt, title=Judge prompt for review model  focusing on visual understanding and reasoning]

[System]

You are a review model tasked with evaluating responses from two assistants to a question about an image.\\

Each assistant has provided an answer based on their analysis of the image.\\

Evaluation Criteria:\\
- Captioning: Evaluate the ability to generate precise descriptions of image elements.\\
- Reasoning: Assess logical consistency and coherence in explanations and conclusions.\\
- Grounding: Evaluate accurate object localization within the image.\\
- Relationship: Assess the understanding of relationships and interactions between subjects in the image.\\
- Focus exclusively on the depth of visual understanding and the quality of reasoning (as described above) in each response. Do not evaluate based on correctness.\\

Evaluation Format:\\
- Compare the two responses impartially. Ignore the order of presentation and the length of the responses. Do not favor any specific assistant based on their name.\\

- Conclude your evaluation by using the following format:\\
  - [[A]] if assistant A's response demonstrates better visual understanding and reasoning,\\
  - [[B]] if assistant B's response demonstrates better visual understanding and reasoning,\\
  - [[C]] if it is a tie.\\

[User Question]\\
\{question\}\\

[The Start of Assistant A's Response]\\
\{Answer\_a\}

[The End of Assistant A's Response]\\

[The Start of Assistant B's Response]\\
\{Answer\_b\}

[The End of Assistant A's Response]\\

[Task]\\
Based on the image, question, and two responses provided, and following the criteria above, determine which assistant provided a better answer focusing solely on visual understanding and reasoning. Use the specified format for your final verdict.

\end{tcolorbox}
}
\end{figure}

\begin{figure*}[ht]
\centering
\resizebox{\textwidth}{!}{
\begin{tcolorbox}[prompt, title=Judge prompt for review model  focusing on correctness]

[System]\\

You are a review model tasked with evaluating responses from two assistants to a question about an image. Each assistant has provided an answer based on their interpretation of the image.\\

Your evaluation is strictly objective, focusing only on which response is correct. Please proceed as follows:\\

1. Assess if Assistant A's response is correct.\\
2. Assess if Assistant B's response is correct.\\
3. Compare the correctness of both responses:\\
   - If only one assistant provides a correct answer, output "[[A]]" if Assistant A is correct, or "[[B]]" if Assistant B is correct.\\
   - If both answers are correct or both are incorrect, output "[[C]]" to indicate a tie.\\

Avoid considering subjective factors such as response quality, detail, or reasoning process. Base your decision solely on the correctness of the answers.\\

[User Question]\\
\{question\}\\

[The Start of Assistant A's Response]\\
\{Answer\_a\}

[The End of Assistant A's Response]\\

[The Start of Assistant B's Response]\\
{Answer\_b}

[The End of Assistant B's Response]\\

[Task]\\

Based solely on the correctness of the two responses, determine which assistant answered the question accurately. Use the specified format for your final verdict.

\end{tcolorbox}
}
\end{figure*}

\begin{figure*}
\centering
\resizebox{\textwidth}{!}{
\begin{tcolorbox}[prompt, title=Answer generation prompt for candidate model]

[System]

Please act as an image-understanding expert to solve the problem based on the provided image.\\

First, analyze the provided image in detail, focusing on its overall theme and key elements.\\

Then, outline your reasoning process step by step, considering how each detail contributes to your understanding of the image.\\

Finally, provide a clear and accurate answer to the user's question based on your analysis. Let's think step by step.\\

[User Question]\\
\{question\} \\

Once you've completed your reasoning, pick one choice from the options. Output the final answer in the format: "[[X]]" where X is the selected option.

\end{tcolorbox}
}
\end{figure*}

\clearpage

\end{document}